\providecommand\IfDocumentMetadataT[1]{}

\documentclass[acmsmall,screen,nonacm]{acmart}

\usepackage{graphicx}

\usepackage{amsmath}
\usepackage{algorithmic}
\usepackage{graphicx}
\usepackage{textcomp}
\usepackage{xcolor}
\usepackage{setspace}
\usepackage{enumitem}
\usepackage{url}
\usepackage{graphics}
\usepackage{tikz}
\usepackage{multirow}
\usepackage{graphicx}
\usepackage{enumitem}
\usepackage{tabularx}
\usepackage{capt-of}
\usepackage{multirow}
\usepackage{amsthm}
\usepackage{listings}
\usepackage{float}
\usepackage{comment}
\usepackage{multirow}
\usepackage{graphicx}
\usepackage{multirow}
\usepackage{makecell}
\usepackage{amsmath}
\usepackage[ruled,lined,boxed,linesnumbered]{algorithm2e}
\usepackage{tikz}
\usetikzlibrary{calc}
\usepackage{multirow}
\usepackage{url}
\usepackage{textcomp}
\usetikzlibrary{shapes, arrows, positioning}
\usepackage{comment}

\usepackage{float}
\usepackage{listings}
\usepackage{makecell}
\usepackage{tabularx}
\usepackage{amsthm}
\usepackage{enumitem}
\usepackage{algorithm2e}

\newcommand{\sk}[1]{\textcolor{black}{ #1}}
\definecolor{instructionviolet}{RGB}{102,51,153}
\definecolor{oratcolor}{RGB}{120,120,120}
\definecolor{son35color}{RGB}{160,160,160}
\definecolor{son46color}{RGB}{31,119,180}
\definecolor{kimicolor}{RGB}{23,162,184}
\definecolor{nosrchcolor}{RGB}{66,135,245}
\definecolor{searchcolor}{RGB}{0,92,175}
\DeclareRobustCommand{\added}[1]{#1}
\newenvironment{addedblock}{\begingroup\color{blue}}{\endgroup}

\SetKwInput{KwOutput}{Output}
\SetKwInput{KwInput}{Input} 

\graphicspath{ {figs/} }    
\setlist[itemize]{noitemsep, topsep=0pt, leftmargin=*}
\usepackage{amsmath}
\usepackage{mathtools}
\usepackage{amsthm}

\usepackage[capitalize,noabbrev]{cleveref}

\theoremstyle{plain}

\theoremstyle{definition}

\theoremstyle{remark}

\newcounter{reviewercount}
\newcounter{commentcount}

\newcommand{\aeditor}%
  {\bigskip\noindent {\bf COMMENTS OF THE ASSOCIATE EDITOR}%
  \setcounter{commentcount}{0}\par 
}

\begin{document}
\raggedbottom

\title{ORFS-agent: Tool-Using Agents for Chip Design Optimization}
\author{Amur Ghose}
\email{aghose@ucsd.edu}
\author{Andrew B. Kahng}
\email{abk@ucsd.edu}
\orcid{0000-0002-4490-5018}
\author{Sayak Kundu}
\email{sakundu@ucsd.edu}
\orcid{0000-0002-8077-1328}
\author{Zhiang Wang}
\email{zhw033@ucsd.edu}
\orcid{0000-0002-6669-9702}
\affiliation{
    \institution{University of California, San Diego}
    \city{La Jolla}
    \state{California}
    \country{USA}
}
\makeatletter
\@ifundefined{authorsaddresses}{}{\authorsaddresses{}}
\makeatother

\begin{abstract}
Machine learning has been widely used to optimize
complex engineering workflows across numerous domains.
In integrated circuit design, modern flows
(e.g., register-transfer level to physical layout)
involve extensive configuration via thousands of
parameters, and small changes can have large
downstream impacts on design performance, power,
and area.
Recent advances in Large Language Models (LLMs) offer
new opportunities for learning and reasoning within
such high-dimensional optimization tasks.
In this work, we introduce \emph{ORFS-agent}, an
LLM-based iterative optimization agent that automates
parameter tuning in an open-source hardware design
flow.
\emph{ORFS-agent} adaptively explores parameter
configurations, demonstrating improvements over
standard Bayesian optimization approaches in terms of
resource efficiency and final design metrics.
\added{Across six benchmarks on ASAP7 and SKY130HD,
thinking-model backends
(Sonnet~4.6 \cite{anthropicClaude4} and
Kimi~K2.5 \cite{kimiK25}) improve the geometric-mean
normalized wirelength, effective clock period, and
co-optimization objectives by up to \(1.0\%\),
\(1.3\%\), and \(2.7\%\) over OR-AutoTuner while
using \(40\%\) fewer iterations; the open-weight
Kimi~K2.5 remains within \(0.24\%\) of Sonnet~4.6,
enabling private deployment. Relative to the earlier
Sonnet~3.5 backend, these thinking models improve the
same objectives by up to \(7.5\%\), \(3.1\%\), and
\(4.0\%\). Optional retrieval tools accelerate early
convergence but do not improve final endpoints.}
By following natural language objectives to trade off
certain metrics for others, \emph{ORFS-agent}
demonstrates a flexible and interpretable framework
for multi-objective \added{and constrained}
optimization. Crucially, \emph{ORFS-agent} is modular
and model-agnostic, and can be plugged into any
frontier LLM without any further fine-tuning.
\added{We also report checkpoint-aligned trajectories
and reasoning summaries that document the agent's
decision process.}
\end{abstract}


\maketitle

\section{Introduction}
\label{sec:intro}

Large Language Models (LLMs) have reshaped AI, excelling at natural-language generation, question answering, and zero-/few-shot learning \cite{brown2020language}. Their emergent reasoning spans mathematical problem solving, code generation, and multi-step orchestration via agent frameworks \cite{drori2022neural}. LLMs integrate into agent systems \cite{wang2024survey}, combining API calls, predefined functions (“tool use”), and context windows to meet language-defined goals.  Agents already serve a range of fields from healthcare to voice-controlled computing.

Research on LLMs has seen 
 advances in neural architecture search \cite{elsken2019neural} \cite{liu2018darts},
combinatorial optimization \cite{bengio2021machine} \cite{mazyavkina2021reinforcement},
and heuristic selection \cite{burke2013hyper} \cite{kotthoff2016algorithm}, all of which are 
relevant to chip design tasks. Chip design relies on a complex 
EDA flow that turns high-level hardware descriptions into manufacturable 
layouts. Each stage exposes hundreds or thousands of hyperparameters (timing constraints,
placement strategies, routing heuristics, etc.) whose joint settings strongly affect 
power, performance, and area (PPA). At advanced nodes, design complexity and the design flow parameter space balloon further \cite{AgnesinaCL23}.

\textbf{Problem setting.}
We target \textbf{LLM-driven black-box optimization} across varied objectives, 
circuits, and technology nodes.
In ML terms, this is \textbf{context-based} Bayesian optimization: LLM agents 
exploit metadata (e.g., run logs) as heuristics. Traditional Bayesian optimization 
(BO) can tune EDA parameters well, but pure black-box methods miss domain knowledge and 
need explicit objectives. By contrast, 
LLMs have broad “world knowledge” and can use chain-of-thought to iteratively propose and 
refine parameters \cite{Zhong2023llm4eda} \cite{Wu2024chateda}.

\textbf{LLM agents} interface with chip design tools, parse intermediate results, 
call domain APIs, and adjust flow parameters in a human-like loop.
They can boost the efficiency of heuristic frameworks (e.g., BO).

\textbf{Relation to prior work: fine-tuning and model-agnostic approaches.}
Many recent efforts apply domain-specific fine-tuning. 
ChatEDA~\cite{Wu2024chateda}, ChipNeMo~\cite{liu2024chipnemo}, and 
JARVIS~\cite{pasandi2025jarvis} curate EDA-centric corpora, retrain or 
instruction-tune a base model, and then self-host the resulting checkpoint; this entails recurring 
data-collection, training, and maintenance costs whenever a stronger foundation 
model is released. ChatEDA, for instance, fine-tunes an open-source (or API-tunable) 
model on Q-A examples. Each new SOTA model demands another fine-tune: closed models 
(e.g., Claude) seldom allow this, and open models lag the frontier. Fine-tuning plus 
lifecycle management and hosting may cost USD \$10k, and the maintainer pays for 
inference.

Other methods avoid retraining, and instead wrap or retrieve around 
frozen SOTA models. EDA-Copilot~\cite{edacopilot2025}, Ask-EDA~\cite{shi2024askeda}, 
OpenROAD-Assistant~\cite{sharma2024openroad}, RAG for EDA-QA~\cite{Pu2024CustomizedRAG}
and ORAssistant~\cite{kaintura2024orassistant} combine retrieval-augmented 
generation with off-the-shelf models for tool-usage questions, while 
surveys ~\cite{Zhong2023llm4eda}~\cite{yu2024futuremirage}~\cite{rai2025enhancingeda} 
highlight the draw of lightweight, model-agnostic agents.

Our agent follows this latter paradigm: it wraps whatever 
SOTA model is available, 
with zero extra training cost; \added{as an example, our full experiments can run 
on Grok-Mini for less than USD \$10.} Further, while previous
agents such as ChatEDA build general Q-A support for engineers,
our agent focuses solely on optimizing flow parameters. \added{Our workflow is validated both on high-end proprietary models (Sonnet-class, Anthropic) and OSS models (Kimi-class, Moonshot AI).}

\added{Reasoning-model backends and bounded retrieval tooling materially change the
best operating point of such an agent. In the final reported configuration,
Sonnet~4.6 reaches geometric-mean normalized objectives of \(0.801\), \(0.839\),
and \(0.855\) for \(WL\)-only, \(ECP\)-only, and co-optimization, respectively;
Kimi~K2.5 reaches \(0.802\), \(0.841\), and \(0.854\). These values improve on both
OR-AutoTuner and the earlier Sonnet~3.5 backend, while also clarifying the
limited but measurable role of retrieval in low-iteration convergence.}
\added{Figure~\ref{fig:front_summary_plots} summarizes these updated results
upfront: the thinking-model backends beat OR-AutoTuner on geometric mean with a
\(40\%\) smaller iteration budget, and retrieval primarily changes the
\(200\)-iteration operating point rather than the final endpoint.}

\textbf{Our contributions.}
We develop \textbf{ORFS-agent}, an LLM-driven autotuner for an open-source 
chip-design flow that can surpass Bayesian optimization using fewer flow runs. 
Our main contributions are:

\begin{itemize}[leftmargin=*]
\item We embed ORFS-agent in OpenROAD-flow-scripts
(ORFS) \cite{ORFS}. It schedules parallel ORFS runs, reads
intermediate metrics (wirelength, timing, power), and
iteratively refines parameters to optimize post-route
results. Leveraging ORFS’s METRICS2.1
logging \cite{jung2021metrics}, the agent skips bad
configurations and gives robust, efficient
\sk{chip design} optimization. Modularity allows
frictionless switching of models.

\item ORFS-agent obeys natural-language instructions to weight or constrain design 
metrics (e.g., to favor timing over wirelength) and is able to leverage existing BO libraries inside 
its tool-use loop.

\item \added{On ASAP7 and SKY130HD (three circuits each), ORFS-agent with
Sonnet~4.6 improves the geometric-mean normalized \(WL\), \(ECP\), and
co-optimization objectives by \(1.0\%\), \(1.3\%\), and \(2.6\%\), respectively,
over OR-AutoTuner while using \(40\%\) fewer iterations; Kimi~K2.5 achieves
\(0.9\%\), \(1.1\%\), and \(2.7\%\) improvements under the same budget.}
\item \added{Relative to the earlier Sonnet~3.5 backend, Sonnet~4.6 improves the
same geometric-mean objectives by \(7.5\%\), \(3.1\%\), and \(3.9\%\), while
Kimi~K2.5 improves them by \(7.4\%\), \(2.9\%\), and \(4.0\%\). Kimi stays within
\(0.24\%\) of Sonnet~4.6 on all three geometric-mean objectives and slightly
improves the co-optimization average, while enabling open-weight deployment.}
\item \added{A retrieval study built around Brave web search and OpenAlex scholarly
lookup, showing that retrieval improves five of six circuit-level
\(200\)-iteration co-optimization checkpoints and lowers the geometric-mean
co-optimization objective by \(0.35\%\) to \(0.36\%\), but degrades all six
\(600\)-iteration co-optimization endpoints by \(0.57\%\) to \(0.60\%\).}
\item \added{Checkpoint-aligned ASAP7-IBEX trajectory and reasoning analyses, plus
Kimi~K2.5 reruns of the auxiliary robustness experiments (constrained
optimization, statistical significance, prompt sensitivity, and obfuscation),
so this journal submission reports both the optimization outcomes and the
decision process that produced them.}
\end{itemize}

\begin{figure*}
\centering
\includegraphics[width=0.78\textwidth]{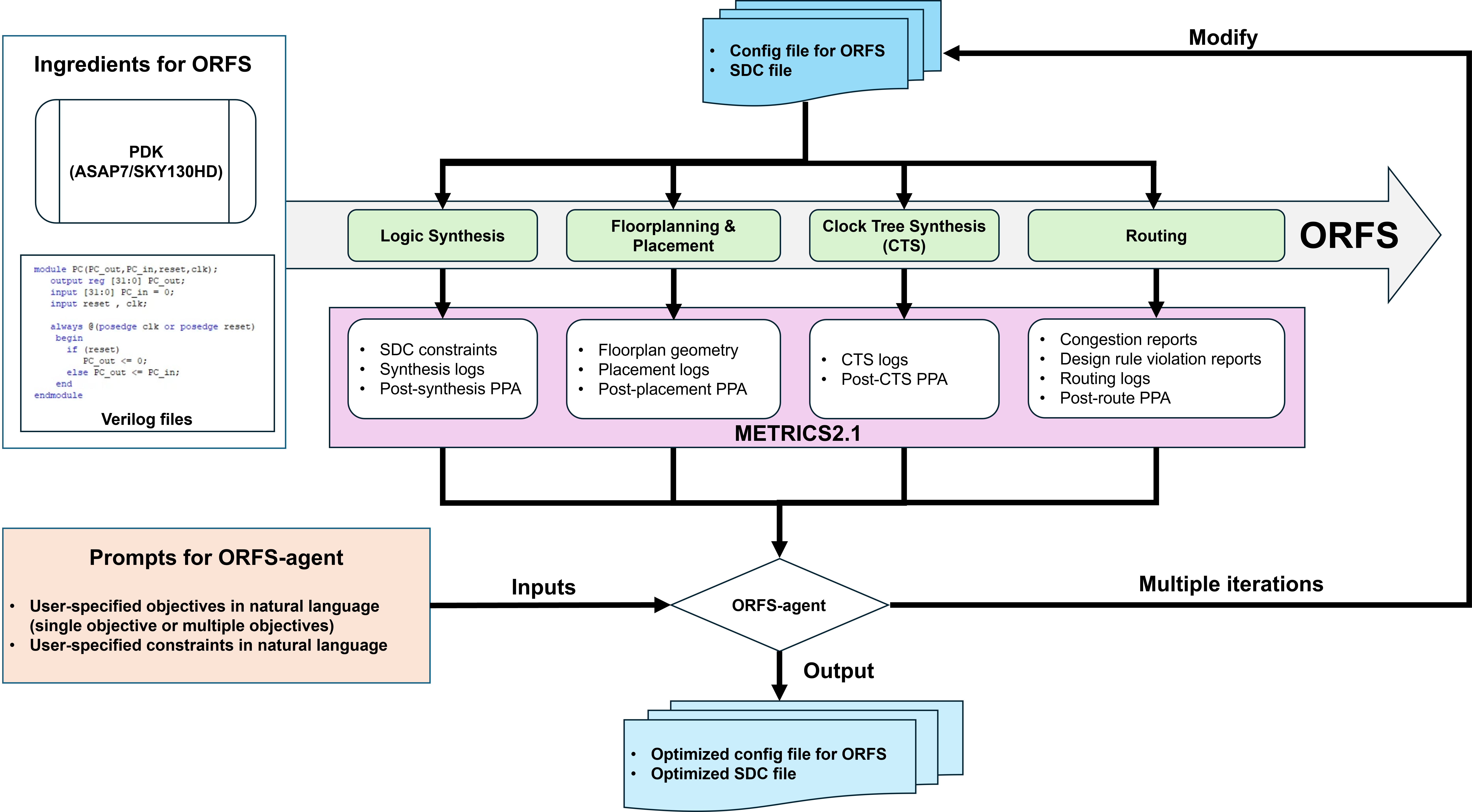}
\caption{ORFS-agent integrated with OpenROAD-flow-scripts (ORFS); metrics gathered via METRICS2.1~\cite{jung2021metrics}.}
\Description{System diagram of ORFS-agent integrated with OpenROAD-flow-scripts: an LLM proposes flow parameters, ORFS runs generate metrics logs, and results feed back into the next iteration.}
\label{fig:OpenROADoverview}
\end{figure*}

\begin{addedblock}
\begin{figure*}[t]
\centering
\includegraphics[width=0.97\textwidth]{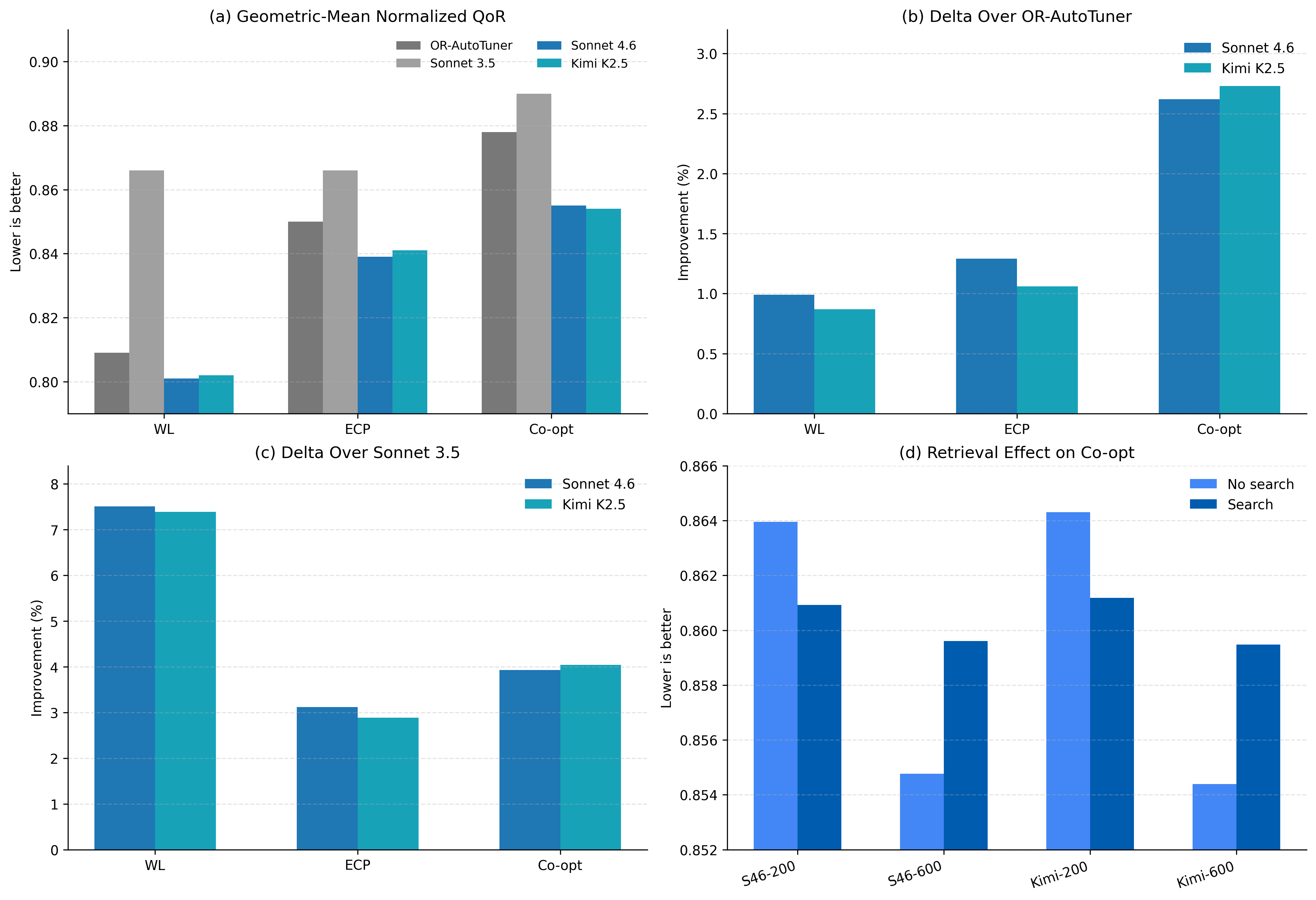}
\caption{\added{Front-summary plots for the new-model and retrieval results.
Panel (a) shows the geometric-mean normalized \(WL\), \(ECP\), and co-optimization
objectives across the six reported benchmarks. Panels (b) and (c) convert those
values into percentage improvements over OR-AutoTuner and the earlier
Sonnet~3.5 backend. Panel (d) shows that retrieval helps the \(200\)-iteration
co-optimization checkpoint for both Sonnet~4.6 and Kimi~K2.5, but regresses the
\(600\)-iteration endpoint.}}
\Description{Four compact bar charts summarizing the new-model and retrieval results: geometric-mean normalized quality of result across models, percentage improvements over OR-AutoTuner, percentage improvements over the earlier Sonnet 3.5 backend, and the effect of retrieval on the co-optimization geometric mean at 200 and 600 iterations.}
\label{fig:front_summary_plots}
\end{figure*}
\end{addedblock}

In the following, Section~\ref{sec:related} reviews BO and recent 
LLM-EDA work. Section~\ref{sec:method} gives details of ORFS-agent -- prompts, data, and iterations. 
Section~\ref{sec:exp} gives experiments. Section~\ref{sec:conclusion} concludes and 
sketches future work. 
Our code is open-sourced at~\cite{orfsagent_code}.

\section{Related Work} 
\label{sec:related}

We categorize related work into three areas: Open-Source EDA, Bayesian Optimization, and LLMs for
Optimization in EDA. Section~\ref{sec:orfsvscomm} gives
additional discussion of EDA flow and OpenROAD.

\subsection{Open-Source EDA and OpenROAD as an ML Testbed}

We target LLM/ML EDA research based on the permissively
open-sourced \textbf{OpenROAD} \cite{ajayi2019OpenROAD}, whose
code, logs and data can be made public to enable reproducible training 
and evaluation.  Tcl/Python APIs expose every flow stage for capture of
live metrics and debug, which is vital  for ML agents.  
With its RTL-to-GDSII scope, transparency and active community, 
OpenROAD provides a robust testbed. {\bf ORFS} wraps OpenROAD 
in editable JSON/Tcl for large-scale experiments \cite{ORFS}.

\subsection{Bayesian Optimization in Flow Tuning}

Bayesian optimization (BO) is a widely used approach for parameter autotuning
in EDA flows~\cite{jung2021metrics}, 
including in commercial tools such as 
Synopsys DSO.ai \cite{DSO} and 
Cadence Cerebrus \cite{Cerebrus}.
BO is efficient in terms of samples and solution space search, 
but has several drawbacks.
\begin{itemize}[leftmargin=*]
\item \emph{Contextual and knowledge limitations.} BO frameworks typically do not 
incorporate domain knowledge unless manually encoded (e.g., at PDK level). Further, 
they do not adapt from past run outcomes during the search.
    
\item \emph{Explicit objective functions.} BO frameworks require explicitly 
defined objective functions, possibly missing complex or subjective criteria. 
This can hinder effectively balancing PPA in chip design.
 
\item \emph{Scalability challenges.} High-dimensional spaces and large designs can 
combinatorially overwhelm traditional BO approaches~\cite{Xu2008}. 

\end{itemize} 
We seek flexible, context-aware and data-driven
approaches to parameter tuning. Large Language Models offer a promising foundation.

\subsection{Large Language Models in EDA Optimization}

LLMs such as GPT \cite{brown2020language} absorb web- and 
paper-scale corpora, giving them broad technical knowledge. They parse design logs, 
infer context, and suggest EDA 
optimizations~\cite{wei2022chain} \cite{Zhong2023llm4eda} \cite{Wu2024chateda}.

\noindent\textbf{Contextual knowledge and adaptivity.}
Unlike black-box Bayesian optimization, LLMs ingest circuit data, retrieve 
literature, and adjust strategies on the fly. They read partial metrics 
(e.g., post-clock tree synthesis (-CTS) timing) and learn user preferences without explicit 
objectives~\cite{ouyang2022training}.

\noindent\textbf{Recent progress.}
Advisor systems LLM4EDA~\cite{Zhong2023llm4eda} and ChatEDA~\cite{Wu2024chateda}
guide placement and timing fixes. Startups
Silogy~\cite{silogyYC25}, 
Silimate~\cite{silimate25},
Atopile~\cite{atopile25}, Diode 
Computers~\cite{diodeYC25} and ChipAgents~\cite{chipagents25} extend AI across design 
tasks. New methods such as LLAMBO~\cite{Liu2024llambo}~\cite{LLAMBO} 
and works presented at ICLAD~\cite{ladSymposium25} and MLCAD (including work
in~\cite{orfsagent_mlcad25})
reinforce the trend. Multi-fidelity BO and 
meta-surrogates~\cite{falkner2018bohb}~\cite{kandasamy2017mfbo}~\cite{shu2024metabo}
improve sampling efficiency yet still treat flows as opaque; by contrast,
LLMs promise context-aware tuning. 
\added{Complementary to end-to-end optimization agents, retrieval-augmented EDA
assistants such as EDA-Copilot~\cite{edacopilot2025}, Ask-EDA~\cite{shi2024askeda},
OpenROAD-Assistant~\cite{sharma2024openroad}, RAG for EDA-QA~\cite{Pu2024CustomizedRAG},
and ORAssistant~\cite{kaintura2024orassistant} show that tool and retrieval stacks can
rapidly bootstrap domain context for downstream decision-making.}
Large-scale industrial adoption of LLMs for EDA can be seen, e.g., NVIDIA’s MARCO 
framework shows graph-based multi-agent task solving for design 
flows \cite{nvidia2025marco}.
Open problems in 
LLM-based hardware verification are mapped in~\cite{wu2024hardwareLLM}.

\subsection{Working with LLMs and Agent Frameworks}

\noindent 
LLMs require scaffolding beyond raw prompts.  Techniques include:
\begin{itemize}[leftmargin=*]
\item \emph{Prompt engineering} crafts system/user messages for clarity.  
\item \emph{Chain-of-thought} reasoning~\cite{wei2022chain} elicits stepwise 
logic.  
\item \emph{Retrieval-augmented generation}~\cite{lewis2020retrieval} injects 
external documents.  
\item \emph{Tool-use/agent 
frameworks}~\cite{qin2023toolllm}~\cite{ge2023openagent}~\cite{Patil2023Gorilla}
let an LLM call 
shells or solvers, parse error logs~\cite{Jimenez2023swebench}, and orchestrate 
full flows~\cite{Khattab2023}.  
\end{itemize}
\noindent
These methods enable adaptive, metric-driven tuning.

\noindent\textbf{LLM agents for optimization.}
Agents excel at ill-specified goals~\cite{Yang2023opro}, prompt 
search~\cite{guo2023connecting}, hyperparameter tuning~\cite{liu2024large}, 
autonomous research~\cite{ifargan2025autonomous}, and benchmark 
challenges~\cite{huang2023benchmarking}. 
EE Times has forecast the advent of collaborating AI agents 
for chip design by 2025 \cite{eetimes2024agents}.

\noindent\textbf{Coupling with Bayesian optimization.}
LLM-based agents can parse PPAC metrics, narrow BO search spaces, prune bad 
configs, or switch heuristics. Blending BO with symbolic 
reasoning~\cite{pan2023logic} \cite{trinh2024solving} yields targeted, adaptive flows. 
\textbf{ORFS-agent} builds on this foundation.

\noindent\textbf{More modern advances.}
OPRO~\cite{Yang2023opro} treats the model itself as the optimizer. 
LLAMBO warm-starts BO with zero-shot hints \cite{Liu2024llambo}. 
ADO-LLM fuses analog priors with BO loops \cite{Yin2024ado}. 
ChatEDA \cite{Wu2024chateda} and the newer EDAid \cite{wu2025edaid} systems wire agents straight into complete RTL-to-GDSII 
flows.
RL efforts such as AutoDMP \cite{Agnesina2023autodmp} \cite{cheng2023assessment} 
and Google’s graph-placement policy \cite{Mirhoseini2021graph} show tractability of 
end-to-end layout.

\noindent\textbf{Tool-centric agent stacks.}
Browser agents such as WebArena~\cite{Zhou2024WebArena} surf, click, and scrape constraints for downstream 
solvers.  
Terminal aides such as ShellGPT~\cite{ShellGPT2023} and Devin~\cite{Devin2024} run \texttt{make}, \texttt{git}, and 
\texttt{yosys} on command, closing the loop between code and 
silicon.  
API routers such as ToolLLM~\cite{qin2023toolllm}, Gorilla~\cite{Patil2023Gorilla}, 
and OpenAgents~\cite{Xie2023OpenAgents} map natural-language tasks onto 
thousands of REST calls and 
plugins.  
AutoGPT~\cite{Sung2023AutoGPT}-style self-looping controllers review results and re-issue 
actions, while the AI-Scientist~\cite{Sakana2024AIScientist} pipeline 
shows how the same pattern can automate hypothesis, experiment, and paper generation.  
At code level, AlphaCode 2~\cite{DeepMind2023AlphaCode2} 
beats 85\% of Codeforces competitors; RL-trained VeriSeek~\cite{Liu2024VeriSeek} and retrieval-guided 
logic synthesis agents~\cite{Chowdhury2024RGSynth} trim Verilog area-delay in minutes.  
Together, browser, tool, and terminal hooks turn LLMs into end-to-end optimization engines. 

	Our work explores how these engines might advance the state of the art in heuristics-driven applied optimization domains such as EDA.

\section{Proposed Methodology}
\label{sec:method}

We now introduce the integration framework of our ORFS-agent 
with OpenROAD-flow-scripts (ORFS).
We then present an explanation of a single iteration of the ORFS-agent loop.

\subsection{Overview of ORFS-agent}
\label{sec:single-llm-agent}

ORFS-agent is an end-to-end agent framework 
designed to optimize OpenROAD-based chip design flows 
according to user-specified objectives and constraints.
ORFS-agent leverages an \sk{LLM} capable
of performing internal text-based reasoning and invoking
external \emph{tools} (function-calling).
Our approach supports parallel runs of ORFS, partial metric gathering 
and various optimization strategies.
\added{In this work, we swap the LLM backend between
Sonnet~3.5, Sonnet~4.6 \cite{anthropicClaude4}, and
Kimi~K2.5 \cite{kimiK25} without changing the agent
loop, and we optionally enable bounded
Retrieval tools (\texttt{WebSearch} and
\texttt{ScholarlyLookup}) to bootstrap
parameter semantics early in an optimization run.}
Figure~\ref{fig:OpenROADoverview} shows
the overall framework of ORFS-agent. 

The inputs to ORFS-agent include process design kits (PDKs) for the target 
technology node, 
target circuits described in Verilog files, and user-defined prompts. 
PDKs are sets of files that provide process-related information needed for chip 
implementation.
In this work, we use two technology nodes: SKY130HD \cite{SKY130HD}
and ASAP7 \cite{ASAP7}.
The prompts define the objective function and constraints in natural language.
The objective function in this work is formulated as
\[F = \sum_{i=1}^n \alpha_i f_i, \]
where each \(f_i\) is a design metric (e.g., routed wirelength, worst 
negative slack) 
and each \(\alpha_i\) is a nonnegative real weight. 
This formulation allows optimization of either a single metric or a weighted 
combination of multiple metrics.
Then, ORFS-agent iteratively modifies the configuration file (config.mk) for ORFS 
and SDC files to optimize the specified objectives. 
The ORFS configuration file defines adjustable parameters (knobs) for
floorplanning, placement, clock tree synthesis and routing.
The SDC file specifies timing constraints for the target design, such as clock 
period.
The outputs of ORFS-agent are the optimized configuration file for ORFS and the 
optimized SDC file; together, these optimize the user-specified objectives under 
given constraints.

ORFS-agent relies on an LLM (e.g., GPT, Claude, Llama) that can 
(i) read and modify files and logs; 
(ii) optionally invoke external tools via function calling, i.e., specify
\sk{argument values} for pre-existing functions (e.g., Python functions);
and (iii) propose new parameter values to improve the objective under given 
constraints.
There is a key distinction between direct LLM use and the use of \emph{LLM-based 
agents} with \emph{function-calling} capabilities. 
In the function-calling paradigm, the LLM is given a goal (e.g., ``write a function 
to compute a matrix eigendecomposition'') 
along with a list of \emph{tools}, which are black-box functions the LLM can invoke 
to facilitate task completion.
For example, a tool named ${\mathbf{ISDIAGONALIZABLE}}$ determines whether 
a matrix \( M \) is diagonalizable
and returns a Boolean value. 
The tools used in ORFS-agent fall into the following \added{four} categories. 
\begin{itemize}[leftmargin=*] 
\item \textbf{INSPECT tools} allow the agent to \emph{inspect} the data collected 
so far without manual in-context inspection of every value.
\item \textbf{OPTIMIZE tools} allow the agent to \emph{model} and \emph{optimize} 
the data using models such as Gaussian-process surrogates.
\item \textbf{AGGLOM tools} allow the agent to {\em sub-select} from possible 
hyperparameters returned by an earlier {\textbf{OPTIMIZE}} step.
\item \added{\textbf{RETRIEVAL tools} allow the agent to query external web and
scholarly sources (e.g., OpenROAD documentation, tuning heuristics) in a bounded,
auditable manner.}
\end{itemize}

\noindent
In the basic function-calling agent loop, 
at iteration \( t \) with context \( C_t \), the agent performs the
following up to \texttt{K} times:
\begin{itemize}
\item \textbf{Observe}: Load the ORFS configuration file and SDC file,
execute ORFS, and examine the resulting flow outputs;
\item \textbf{Query}: Call tools from the provided list (up to \( M \) times) to 
gather information and update the context to \( C_{t+1} \); and
\item \textbf{Alter}: Modify ORFS config/SDC files based on \( C_{t+1} \).
\end{itemize}
By iterating through these steps, the function-calling agent incrementally refines 
its solutions by selectively invoking external tools.

\subsection{Overall Iteration Structure with Example and Tool Usage}

\begin{figure}
    \centering
    \includegraphics[width=1\linewidth]{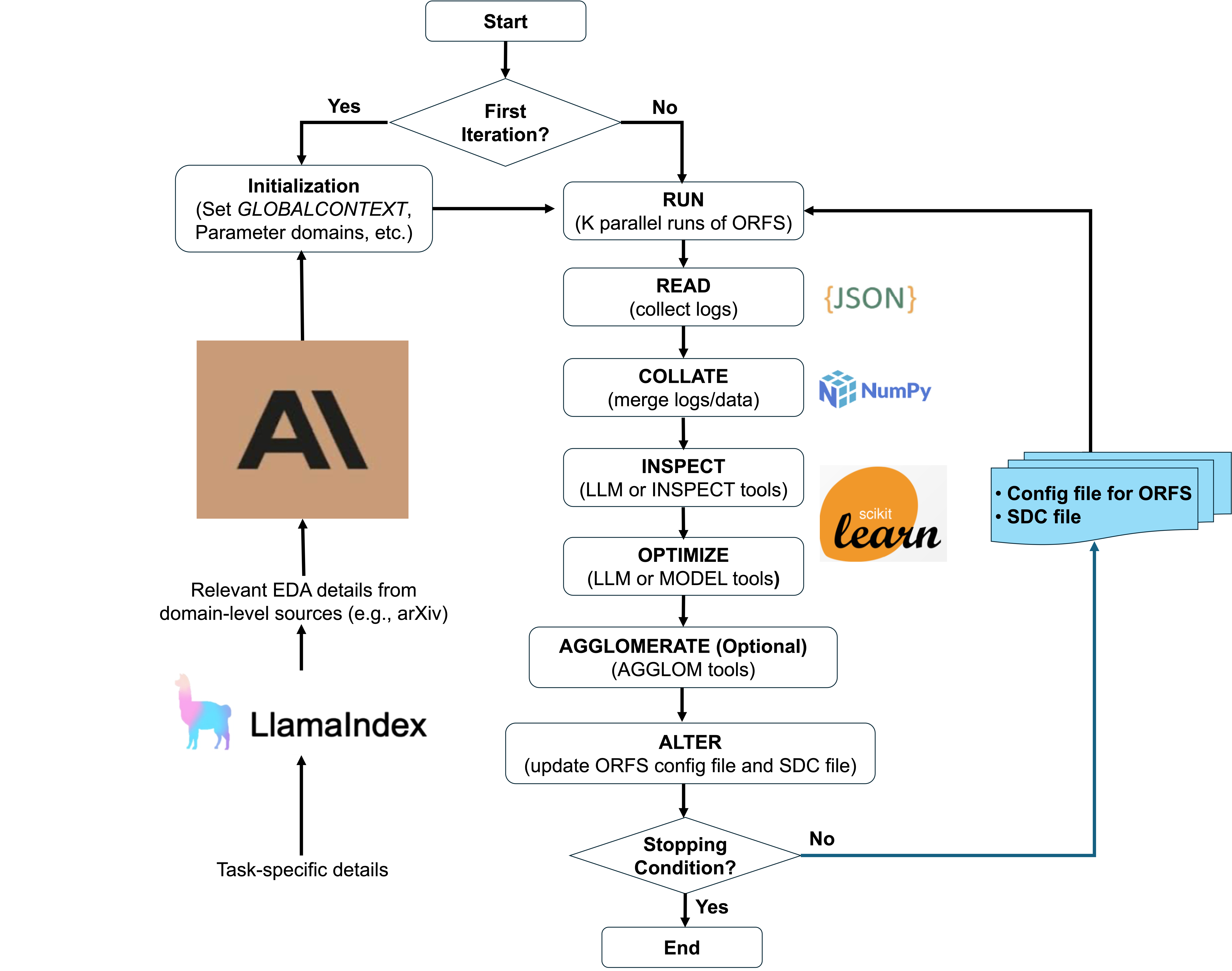}
    \caption{Overall ORFS-agent flowchart. \added{Retrieval tools (\texttt{WebSearch}
    and \texttt{ScholarlyLookup}) may be invoked during INSPECT/OPTIMIZE when enabled,
    without changing the loop structure.}}
    \Description{Flowchart of one ORFS-agent iteration: run K parallel jobs, read and collate logs, inspect results, optimize and optionally agglomerate candidates (with optional retrieval when enabled), then alter parameters for the next batch.}
    \label{fig:ORFS-agent}
\end{figure}

We describe \textbf{one} iteration of the ORFS-agent loop, 
which executes \texttt{K} parallel runs of OpenROAD-flow-scripts, 
with \texttt{K} being provided by the user. 
As shown in Figure~\ref{fig:ORFS-agent}, 
each iteration consists of the following steps: \textbf{RUN}, 
\textbf{READ}, 
\textbf{COLLATE}, 
\textbf{INSPECT}, 
\textbf{OPTIMIZE},
\textbf{AGGLOMERATE} (optional)
and \textbf{ALTER}. 
\added{When enabled, retrieval calls are made within INSPECT/OPTIMIZE and are
bounded per iteration; see Section~\ref{sec:tooldetails}.}

We maintain a \texttt{GLOBALCONTEXT} that collects relevant parameters and 
partial metrics across steps. The initialization of this \texttt{GLOBALCONTEXT} at 
the first iteration is also the LLM prompt; this initialization, along with 
procedures for update between iterations for \texttt{GLOBALCONTEXT}, is separately 
discussed in Section~\ref{sec:methodapp}. Let \texttt{K} denote the number of
parallel 
runs per iteration, and let \texttt{TIMEOUT} specify the maximum runtime for 
each batch. Below, we show an example scenario with
\(\texttt{K}=25\), \(\texttt{TIMEOUT}=30\) minutes and parameters 
(\textbf{Core Util}, \textbf{Clock Period}) that can be integrated, along with 
possible function-calling tools. We next consider the $j^{th}$ iteration and
assume that we target the \textbf{routed} wirelength, taking the \textbf{CTS}
wirelength as a surrogate (recalling that the CTS stage is earlier and thus
likely does not time out).

\noindent
\textbf{Stage 1: RUN} \texttt{K} \textbf{Parallel Jobs.}
    Launch \texttt{K} ORFS runs, each with distinct parameters, until the 
    allotted \texttt{TIMEOUT} expires.
    \begin{itemize}
        \item \emph{Example:} For \(\texttt{K} = 25\) and \(\texttt{TIMEOUT}=30\) 
        minutes, we launch 25 parallel runs, each exploring a different 
        (\textbf{Core Util}, \textbf{Clock Period}) setting. All will run for at 
        most $30$ minutes, with one log file per run.
        \item \emph{Tool use:} This set of runs might be derived 
        from a previous iteration where \textbf{OPTIMIZE} or \textbf{AGGLOM} 
        tools suggest new parameter sets. 
        \item \emph{No tool use:} If tool use is disabled, these parameter files 
        arise from simple in-context operations.
    \end{itemize}

\noindent
\textbf{Stage 2: READ.}
    Gather JSON logs from the files generated by all \textbf{RUN} steps up to this iteration.
    These may contain partial surrogate
    metrics, e.g., from the CTS stage, or ground truth metrics from 
    runs that finished before timeout.
    \begin{itemize}
        \item \emph{Example:} The agent collects $25 \times j$ JSON files, with 
        routed wirelengths (ground truth) or earlier rough CTS wirelengths 
        (surrogate).
        \item \emph{No tools are used here}, so behavior is identical with or
        without tool access.
    \end{itemize}

\noindent    
\textbf{Stage 3: COLLATE.}
    Merge the logs into a single dataset, storing outcomes for all 
    \texttt{K} $\times$ $j$ runs (e.g., a table or array).
    \begin{itemize}
        \item \emph{Example:} Build a unified array with columns 
        (\textbf{Core Util}, \textbf{Clock Period}, CTS wirelength, routed 
        wirelength). The routed wirelength can be absent. If the CTS wirelength is 
        absent, skip the row.
        \item \emph{No tools are used here}, so behavior is identical with or
        without tool access.
    \end{itemize}

\noindent    
\textbf{Stage 4: INSPECT.}
    Analyze the collated data.
    \begin{itemize}
        \item \emph{Example:} For $\leq 25 \times j$ ($\leq$ from skips) instances, 
        find how 
        \textbf{Clock Period} varies against the wirelength (actual target)
        and CTS wirelength (surrogate).
        \item \emph{Tool use:} The agent calls pre-defined \textbf{INSPECT} tools  
        to examine distributions; this allows sidestepping of in-context inspection of  values.        
	        For example,  $\mathbf{InspectDistribution(ARR, X,Y,Y_{surrogate})}$
	        outputs the distributional relation between an input $\mathbf{X}$ and 
	        targets $\mathbf{Y}$ with surrogate targets $\mathbf{Y_{surrogate}}$. 
	        Here we have \textbf{Clock Period} as $\mathbf{X}$, CTS and routed wirelength 
	        as $\mathbf{Y_{surrogate},Y}$ and $\mathbf{ARR}$ from the \textbf{COLLATE} step. 
	        \item \added{\emph{Optional retrieval:} The agent may call
	        RETRIEVAL tools (\texttt{WebSearch} and \texttt{ScholarlyLookup}) to fetch
	        short, cached snippets about parameter semantics or related work. These
	        snippets are capped and treated as untrusted input, and are used to inform
	        the subsequent OPTIMIZE call.}
	        \item \emph{No tool use:} The agent reasons purely in context 
	        using its chain of thought by inspecting the data.
	    \end{itemize}

\noindent
\textbf{Stage 5: OPTIMIZE.}
    Based on \textbf{INSPECT}, propose new parameter sets for the next iteration. 
    \begin{itemize}
        \item \emph{Example:}  The agent calls an \textbf{OPTIMIZE} tool, 
         e.g., a Gaussian process ($\mathbf{GP}$) with the \textbf{COLLATE}-level 
         data, to get $200$ values of candidate \textbf{Clock Period, Core Util} 
         pairs, using a \textbf{Matérn kernel}, and \textbf{Expected Improvement} 
         as search strategy.
        \item \emph{Tool use:} In general, we may have candidate points from $
        \mathbf{GP}(\mathbf{ARR},\mathbf{args},\dots)$  or any other 
        \textbf{OPTIMIZE} tool, obtaining many potential parameter sets. 
        $\mathbf{ARR}$ arises from collate, and the agent decides on which 
        \textbf{OPTIMIZE} tool to use, 
         i.e., whether to use a Gaussian process at all, as well as the choice of 
         $\mathbf{args}$ depending on what it finds at the \textbf{INSPECT} phase. 
         This may include choices over the kernel used, selection algorithm, and so 
         on.
         \item \emph{No tool use:} The agent solely uses its own context to directly
        suggest a set of parameters.
    \end{itemize}

    \noindent
    \textbf{Stage 6: AGGLOMERATE (optional).} If more than \texttt{K} 
        candidates are generated in the \textbf{OPTIMIZE} phase, then reduce 
        to exactly \texttt{K}, 
        e.g., by maximizing diversity or coverage.
        \begin{itemize}
            \item \emph{Example:} Since the  $\mathbf{GP}$ produces $200$ new 
            (\textbf{Core Util}, \textbf{Clock Period}) pairs; we sub-select to 
            25 via a valid \textbf{AGGLOM} tool, 
            $\mathbf{EntropySelect(ARR,quality,points)}$. This function returns 
            $\mathbf{points}$, i.e., $25$ pairs of  (\textbf{Core Util}, 
            \textbf{Clock Period}) that are considered set-level optimal with respect to the 
            provided $\mathbf{quality}$ metric (wirelength improvement) on the same 
            $\mathbf{ARR}$ from \textbf{COLLATE}.
            \item \emph{Tool use:} The agent may use any number of \textbf{AGGLOM} 
            tool choices from the LLM's decisions along with parameters to call 
            them with. The final output is a set of exactly \texttt{K}
            candidate parameter tuples for evaluation.
            \item \emph{No tool use:} Here, the \textbf{OPTIMIZE} case uses 
            structured outputs to yield exactly \texttt{K} sets of parameter 
            choices in context directly, and this \textbf{step is skipped.}
        \end{itemize}

    \noindent
\textbf{Stage 7: ALTER.}
    Update ORFS configuration files with the newly selected parameter sets 
    for the next round of runs.
    \begin{itemize}
        \item \emph{Example:} Generate 25 updated config/SDC files or commands, 
        each having one of the chosen (\textbf{Core Util}, 
        \textbf{Clock Period}) combinations.
        \item \emph{No tools are used here}, so behavior is identical with or
        without tool access.

    \end{itemize}

\noindent
This process repeats until the stopping criteria -- generally, total serial 
iteration count -- are met. 
The outputs of stages \textbf{READ} through \textbf{AGGLOMERATE} enter the context 
window of ORFS-agent in each iteration, but do not necessarily enter the  
\texttt{GLOBALCONTEXT}.
Context management and tool-level details are given in Section~\ref{sec:methodapp}.

\subsection{EDA Flow and OpenROAD}
\label{sec:orfsvscomm}

The EDA flow is highly iterative\sk{, and}
suboptimal outcomes at one step may require going back to earlier steps
for re-optimization (e.g., re-synthesizing the gate-level netlist or adjusting the placement).
Each step generates logs and metrics
that can be integrated into ML pipelines to guide later agent decisions within the flow.

To systematically study LLM- and ML-driven EDA optimizations, we focus on the
\emph{OpenROAD} project~\cite{ajayi2019OpenROAD}. In contrast to commercial EDA
offerings, OpenROAD is fully open-source, thus avoiding copyright restrictions and
enabling access to the internal details of its place-and-route algorithms, log
files, and data structures. This open access allows the researchers to
replicate and extend experiments, train and evaluate new models, and publish
results without proprietary barriers.~\cite{kahng22}

In our work, leveraging OpenROAD allows us to (i) faithfully represent the full
digital implementation flow (floorplanning, placement, clock tree synthesis,
routing, etc.), (ii) integrate with a community-driven research ecosystem, and (iii)
avoid opaque “black-box” workflows that impede effective ML experimentation.
In particular, we are able to explore the potential of LLM-based
EDA agents that can dynamically interpret intermediate results and fine-tune the
flow for better optimization outcomes.

\subsection{Global Context, Initialization, and the Toolbox}
\label{sec:methodapp}

The function-calling variant of \textbf{ORFS-agent}
maintains a global context, denoted \texttt{GLOBALCONTEXT},
that stores the design environment details (PDK,
circuit, task specifics, etc.). Each LLM
call in the pipeline includes this global context, together with the relevant
logs and runtime-extracted data. We initialize \texttt{GLOBALCONTEXT} from
the \emph{prompt} defined below.

LLM calls made within each part of an iteration -- i.e., \textbf{INSPECT}, etc. --
add the \texttt{LOCALCONTEXT} of that particular iteration. However, they are not
necessarily passed on to the \texttt{GLOBALCONTEXT}. As we move from iteration $j$
to iteration $j+1$, \texttt{GLOBALCONTEXT} is modified to only include a small subset of all
the \texttt{LOCALCONTEXT} within the particular iteration $j$.

\subsubsection{Initialization}
At the very first iteration, the prompt is
initialized from the following inputs:
\begin{itemize}
    \item The design platform (e.g., ASAP7).
    \item The circuit under consideration (e.g., AES).
    \item The optimization task (e.g., minimize wirelength ($WL$)).
    \item The input design parameters (e.g., the 12 tunable parameters,
    including core utilization).
    \item The output variable sets, including detailed $WL$ and surrogate
    variables such as CTS $WL$.
    \item The exact output quantity to optimize (e.g., $-WL$ for
    maximization or $WL$ for minimization).
    \item The domains of the inputs, including whether they are Boolean,
    integers with min/max bounds, or continuous values converted to integers
    (e.g., converting float ranges such as 0.01--0.99 to integer ranges such as 1--99,
    and scaling back to original values for \textbf{PARAMETERS} generation).
    \item Suggested ranges for inputs.
\end{itemize}

\subsubsection{Toolbox}

We define \added{four} separate kinds of tools. All of these require
a \textbf{Description}, a set of ideally typed \textbf{arguments}, and a return
object \textbf{Return} with description and type specifications to be efficiently
made a part of LLM function-calling frameworks.
We summarize tool details in
Section~\ref{sec:tooldetails} below.

\textbf{INSPECT Tools.} These tools enable quick analysis of numeric datasets by
computing summary statistics, correlations, outlier indices, or performing
principal component analysis. They do not consume raw data token-by-token but
instead return concise analytical results that aid downstream decision-making.

\textbf{OPTIMIZE Tools.} These tools propose new points (parameter sets)
to explore in an optimization process. Each one uses a different strategy -- ranging
from random sampling and grid-based approaches to more sophisticated algorithms
such as Bayesian optimization or genetic algorithms -- to balance exploration and
exploitation.

\textbf{AGGLOM Tools.} These tools \emph{agglomerate} (reduce or cluster) a large
set of candidate solutions into a smaller or more representative subset. Methods
include selecting the Pareto front for multi-objective problems, maximizing
coverage or diversity, or clustering to find the most central or representative
candidate sets.

\added{\textbf{RETRIEVAL Tools.} These tools retrieve external information on-demand
via web search and scholarly lookup. In our setting, retrieval is bounded and used
to bootstrap parameter semantics and tuning heuristics early in an optimization run
(e.g., by finding OpenROAD/ORFS documentation or design-specific \texttt{config.mk}
artifacts). The retrieval tools we expose are \texttt{WebSearch} via Brave and
\texttt{ScholarlyLookup} via OpenAlex.}

\subsection{Tool Details}
\label{sec:tooldetails}

\subsubsection*{INSPECT Tools}
\textbf{AnalyzeManifold} analyzes the underlying manifold structure using PCA,
TSNE, and MDS. It takes \texttt{X: np.ndarray} and an optional \texttt{config:
Dict}, and returns a compact summary dictionary.

\textbf{AnalyzeLocal} examines local structure using LOF and DBSCAN. It accepts
\texttt{X: np.ndarray} and an optional \texttt{config: Dict}, and returns a summary
dictionary output.

\textbf{Inspect\allowbreak Distribution} analyzes statistical properties of input/output data.
It requires \texttt{X: np.\allowbreak ndarray}, \texttt{Y: np.\allowbreak ndarray}, and optionally
\texttt{Y\_\allowbreak surrogate: np.\allowbreak ndarray}. The output is a compact summary dictionary.

\textbf{InspectStructure} inspects structural properties and gives model
recommendations. Inputs include \texttt{X: np.ndarray}, \texttt{Y: np.ndarray}, and
optionally \texttt{config: Dict}. It returns a compact summary dictionary.

\subsubsection*{OPTIMIZE Tools}
\textbf{Create\allowbreak Model} creates and configures a Gaussian process model. It takes
\texttt{X: np.\allowbreak ndarray}, \texttt{y: np.\allowbreak ndarray}, \texttt{noise\_\allowbreak level: float}, and
\texttt{kernel\_\allowbreak type: str}. It returns a fitted
\texttt{Gaussian\allowbreak Process\allowbreak Regressor} model.

\textbf{CreateKernel} generates a kernel from a given specification. Required
inputs are \texttt{kernel\_spec: str} and \texttt{input\_dim: int}. The output is a
configured Gaussian-process kernel.

\textbf{ExpectedImprovement} calculates the Expected Improvement (EI) acquisition
function. Inputs include \texttt{mu: np.ndarray}, \texttt{std: np.ndarray}, and
\texttt{y\_best: float}. It returns \texttt{ei: np.ndarray}.

\textbf{HandleSurrogate} processes and combines true and surrogate data. It uses
\texttt{X: np.ndarray}, \texttt{y: np.ndarray}, and \texttt{surrogate\_values:
np.ndarray}. It returns a tuple \texttt{(X, y\_combined, uncertainty)}.

\textbf{LatinHypercube} generates Latin Hypercube samples.
The inputs are
\texttt{n\_points: int} and \texttt{n\_dims: int}.
It returns  \texttt{samples:
np.ndarray}.

\subsubsection*{AGGLOM Tools}
\textbf{Select\allowbreak Points} selects points based on a specified method. It accepts
\texttt{X: np.\allowbreak ndarray}, \texttt{quality\_\allowbreak scores: np.\allowbreak ndarray}, \texttt{method: str},
and \texttt{n\_\allowbreak points: int}. It returns \texttt{selected\_\allowbreak indices: np.\allowbreak ndarray}.

\textbf{HybridSelect} combines quality and diversity for point selection. Required
arguments are \texttt{X: np.ndarray}, \texttt{quality\_scores: np.ndarray},
\texttt{distance\_matrix: np.ndarray}, and \texttt{n\_points: int}. It returns
\texttt{selected\_indices: np.ndarray}.

\textbf{Entropy\allowbreak Select} uses entropy-based diversity for selection. It takes
\texttt{X: np.\allowbreak ndarray}, \texttt{quality\_\allowbreak scores: np.\allowbreak ndarray}, and
\texttt{n\_\allowbreak points: int}. It returns \texttt{selected\_\allowbreak indices: np.\allowbreak ndarray}.

\textbf{GraphSelect} applies graph-based diversity metrics for point selection. It
accepts \texttt{X: np.ndarray}, \texttt{quality\_scores: np.ndarray}, and
\texttt{n\_points: int}. It returns  \texttt{selected\_indices: np.ndarray}.

\textbf{Create\allowbreak Quality\allowbreak Scores} computes quality scores from model predictions.
\begin{description}[style=nextline,leftmargin=1.5em,itemsep=0pt,topsep=0pt]
\item[Input] \texttt{X: np.\allowbreak ndarray}
\item[Input] \texttt{y: np.\allowbreak ndarray}
\item[Input] \texttt{model\_\allowbreak predictions: np.\allowbreak ndarray}
\item[Optional input] \texttt{model\_\allowbreak uncertainties: np.\allowbreak ndarray}
\item[Output] \texttt{quality\_\allowbreak scores: np.\allowbreak ndarray}
\end{description}

\subsubsection*{RETRIEVAL Tools}
\textbf{WebSearch} issues a web query (Brave backend) and returns the top results as
a short list of \texttt{\{title, url, snippet\}} records. It takes
\texttt{query: str} and an optional \texttt{config: Dict} (e.g., \texttt{top\_k},
\texttt{site\_filter}), and returns \texttt{results: List[Dict]}.

\textbf{ScholarlyLookup} queries scholarly metadata (OpenAlex backend) to resolve
canonical paper metadata (title, authors, year, venue) and stable identifiers.
It takes \texttt{query: str} and an optional \texttt{config: Dict} (e.g.,
\texttt{top\_k}, \texttt{year\_range}, \texttt{venue\_filter}), and returns
\texttt{works: List[Dict]}.

Each of the tools used here can be implemented using standard Python packages. For
instance, \texttt{scikit-learn, numpy} and related Python packages offer support
for Gaussian processes, t-SNE modeling, and so on. By combining these existing
libraries with domain-specific logic, we create a modular, maintainable toolbox
that reflects the iterative exploration process a human data scientist would
perform -- streamlining tasks such as outlier detection, dimensionality reduction, and
optimization-driven design.

\label{sec:kimi_thinking_search_notes}

\noindent\textbf{Backend and retrieval implementation details for the experiments.}
The following subsubsections record the implementation choices needed to reproduce
and interpret the Sonnet~4.6, Kimi~K2.5, and retrieval results reported later in
Section~\ref{sec:addnl}. They are not separate experiments: they document how we
integrated Kimi~K2.5, what changed when moving from Sonnet~3.5 to later
thinking-capable backends inside the same iterative tool-using loop, and why we
chose Brave (web) and OpenAlex (scholarly metadata).

\subsubsection{Integrating Kimi~K2.5}
\label{subsec:kimi_impl}

ORFS-agent is \emph{backend-agnostic}: a thin adapter normalizes (i) message roles,
(ii) tool/function calling, (iii) strict structured outputs for OPTIMIZE, and (iv)
token/rate-limit budgeting across providers. In practice, Kimi behaves ``OpenAI
compatible'' enough to reuse the OpenAI SDK, but it still has provider-specific
details that affect agent-level pipelines.

\noindent\textbf{Endpoints and compatibility.}
Moonshot exposes an OpenAI-compatible API at \texttt{api.moonshot.ai} (base path
\texttt{/v1}; a regional \texttt{api.moonshot.cn} endpoint is also documented). In our
agent loop, the two most relevant surfaces are \texttt{/v1/chat/completions} and the
file-extraction workflow (\texttt{/v1/files} plus \texttt{/content}) used for
long-context grounding.

\noindent\textbf{Parameter quirks relative to OpenAI ChatCompletions.}
Moonshot documents two practical differences that matter for sampling logic:
(i) the temperature range is $[0,1]$, and (ii) when temperature is set to $0$ (or very
close to $0$), the API only supports returning a single choice, and will error if
\texttt{n>1}. We keep \texttt{n=1} for deterministic settings and emit
multiple proposals within one structured completion.

\noindent\textbf{Tool calling constraints.}
Kimi follows the OpenAI-style \texttt{tools} interface, with tool definitions expressed
as a JSON-Schema subset. Two constraints we explicitly enforce in our adapter are:
(i) tool names must follow the documented naming rule (start with a letter or
underscore; then up to 63 characters from letters, digits, hyphen, or underscore),
and (ii) the number of tools in a single request is bounded (Moonshot documents a
maximum of 128).

\noindent\textbf{Structured outputs: JSON Mode and Partial Mode.}
For strict parsing, Kimi exposes a JSON Mode via
\texttt{response\_format=\{"type":"json\_object"\}}. It supports a
``Partial Mode'' (\texttt{"partial": true}) for output prefilling; Moonshot warns not
to combine Partial Mode with JSON Mode.

\noindent\textbf{Thinking traces.}
For Kimi thinking-enabled models (e.g., \texttt{kimi-k2-\allowbreak thinking} and \texttt{kimi-k2.5}), Moonshot
returns an explicit reasoning trace in a separate \texttt{reasoning\_\allowbreak content} field.
When using the OpenAI SDK, this field is not part of the stock message type and must
be accessed via attribute checks (e.g., \texttt{hasattr}/\texttt{getattr}). In streaming
mode, \texttt{reasoning\_\allowbreak content} appears before \texttt{content}, and Moonshot notes
that the combined token count of \texttt{reasoning\_\allowbreak content} and \texttt{content} is
bounded by \texttt{max\_tokens}; we exploit this property when budgeting output length.

\noindent\textbf{Reproducibility.}
We log model identifiers, prompts, tool invocations/returns, and per-iteration random
seeds so that differences can be attributed to the model, the tools, or
the EDA flow.

\subsubsection{Why ``Thinking'' Models Behave Differently from Earlier Backends}
\label{subsec:thinking_models_impl}

Relative to Sonnet 3.5, thinking-capable models (e.g., Sonnet 4.6) tend to be more
reliable in a tool-using loop and more capable of using intermediate feedback over
long horizons (logs, metrics, and tool summaries).

\noindent\textbf{Fewer retries under strict schemas.}
In the 12-parameter setting, malformed outputs can stall an iteration. In practice,
stronger models reduce schema violations and retries, which matters because retries
increase latency/cost and can perturb the effective context.

\noindent\textbf{Harder objective push (and secondary-metric drift).}
Stronger models can more aggressively optimize the explicitly stated target while
allowing non-target QoRs to drift (e.g., improving $WL$ while worsening $ECP$; see
Section~\ref{subsec:constopt}). If multiple QoRs matter, we found it important to
explicitly encode constraints (acceptable regressions) or to use multi-objective
formulations rather than relying on implicit ``common sense'' tradeoffs.

\noindent\textbf{More appetite for retrieval.}
When search tools are enabled, thinking-capable models often probe external artifacts
more aggressively early in a run. Operationally, we cap search calls per iteration and
cache results to prevent runaway retrieval loops.

\noindent\textbf{Thinking-token math and trace retention.}
Thinking is not ``free'': depending on the provider, it consumes hidden compute,
explicit reasoning tokens, or an exposed trace (e.g., Kimi's \texttt{reasoning\_\allowbreak content})
that competes with the final answer under \texttt{max\_tokens}. Exposed traces can be
useful for debugging and self-correction, but in a \emph{serial} optimization loop they
can also be actively harmful: retaining even a few hundred trace tokens per iteration
makes the prompt footprint grow roughly linearly with iteration count and can crowd
out the highest-value signals (fresh metrics, the newest logs, and the current
candidate pool).

We therefore treat thinking traces as \emph{observability}, not memory: we log them
for audit/debugging, but we do not feed them back verbatim. Instead, we keep a short
structured ``decision summary'' per iteration, and we budget \texttt{max\_tokens} so a
trace (when exposed) cannot crowd out the structured proposal required by OPTIMIZE.
\subsubsection{Search and Scholarly Providers: Tradeoffs and Rationale}
\label{subsec:search_providers_impl}

ORFS-agent optionally uses two retrieval tools (Section~\ref{sec:addnl}): general web
search (for documentation and public repository artifacts) and scholarly metadata
lookup (for citation resolution and metadata hygiene). Retrieval helps bootstrap
domain knowledge (e.g., ``what does this OpenROAD variable control?''), but introduces
retrieval bias, prompt-injection surfaces, and less reproducible ranking.

\noindent\textbf{Why Brave + OpenAlex in particular.}
We chose Brave for web search and OpenAlex for scholarly lookup because both are
API-first and easy to cache and log: Brave is effective for navigational queries
against docs/repos, and OpenAlex provides stable identifiers and structured filters
that support deduplication and reproducible paper lookup.

\begin{table}[!t]
\centering
\footnotesize
\setlength{\tabcolsep}{4pt}
\renewcommand{\arraystretch}{1.1}
\begin{tabularx}{\columnwidth}{|l|X|}
\hline
\textbf{Provider} & \textbf{When it helps (and caveats)} \\
\hline
Brave (web) &
API-first general web index; strong for docs/repos and targeted artifact lookup; easy to cache. Caveat: ranking is not stable over time; retrieved pages can contain prompt-injection text. \\
\hline
Bing / Google CSE (web) &
Very broad coverage and often strong navigational results. Caveat: quotas/terms can be restrictive; ranking is less reproducible and can shift over time. \\
\hline
Aggregator wrappers (web) &
Unified interface across engines; fast to integrate. Caveat: adds an extra dependency layer and can obscure provenance. \\
\hline
Summary APIs (web + LLM) &
Produce short ``LLM-ready'' snippets and reduce context bloat. Caveat: adds a second model in the loop; harder to attribute evidence precisely. \\
\hline
OpenAlex (scholarly) &
Open metadata graph with stable IDs and structured filters; simple caching/dedup. Caveat: primarily metadata (not full text); coverage varies by venue. \\
\hline
Semantic Scholar (scholarly) &
Convenient fields for authors/citations and often strong CS/ML coverage. Caveat: rate limits can constrain long loops; coverage is uneven across fields. \\
\hline
Crossref / DOI registries (metadata) &
Useful for resolving DOIs and canonical citation strings. Caveat: weak for discovery; metadata depends on publisher deposits. \\
\hline
arXiv (preprints) &
Good for rapid access to preprints and versioning. Caveat: limited scope and not a general scholarly index. \\
\hline
\end{tabularx}
\caption{\added{Representative tradeoffs of retrieval providers considered for ORFS-agent. Our deployed combination uses Brave for general web discovery and OpenAlex for scholarly metadata.}}
\label{tab:retrieval_providers_tradeoffs}
\end{table}

\noindent\textbf{Practical mitigations.}
We treat retrieved text as untrusted input, cap retrieval calls per iteration, keep
inserted snippets short, and cache query$\rightarrow$result mappings to improve
reproducibility. In addition, we primarily use retrieval for \emph{concept
clarification} (definitions, parameter meanings, prior work) rather than for injecting
large volumes of external text into context.

\subsubsection{Retrieval Integration Details - Querying, Budgets, and Provenance}
\label{subsec:retrieval_integration_impl}
Queries are synthesized from run context (design, PDK, objective) and log-derived
signals, normalized to stable cache keys, and capped per iteration. We store raw
provider responses (JSON) alongside iteration logs so that any retrieved claim can be
traced to a provider response and reruns can reuse cached results.

\noindent\textbf{Snippet selection and sanitization.}
We only insert short snippets (titles, URLs, small passages) and bound the total
retrieval payload per iteration. Retrieved text is treated as untrusted data: we strip
markup, drop obviously imperative patterns, and keep retrieval content isolated from
system/agent instructions so it cannot override objectives or constraints.

\subsubsection{Self-Referential Retrieval}
\label{subsec:self_ref_retrieval}

With retrieval enabled, a sufficiently literal model will occasionally rediscover the
ORFS-agent paper itself on arXiv when asked to search for OpenROAD tuning or flow
variables. In effect, the agent uses the search APIs to find the document that
describes the agent.

This observation also reinforces retrieval hygiene: self-referential
hits do not automatically imply cheating or data leakage, yet they can bias an agent
toward highly retrievable sources (including its own artifacts). For this reason, we
keep retrieval optional, bounded, and logged, and we rely primarily on current-run
metrics/logs for decision-making.

\section{Experiments and Results}
\label{sec:exp}

 Our experiments use six design instances
comprising three circuits (IBEX, AES and JPEG) in two technology 
nodes (SKY130HD and ASAP7).
All experiments are performed on a Google Cloud Platform (GCP) 
virtual machine with 112 virtual CPUs (C2D AMD Milan) and 
220GB RAM. \added{All experiments use the ce8d36a commit of the
OpenROAD-flow-scripts (ORFS) repository~\cite{ORFS}, together with the
corresponding versions of OpenROAD, KLayout, Yosys, and related
dependencies. The Sonnet~3.5 results reused from~\cite{orfsagent_mlcad25}
were obtained in the same ORFS environment. For the matched cross-model
comparisons against that reference, we also keep the same serial-iteration
checkpoints and the same \(375\)/\(600\)-iteration budgets.}

\added{Section~\ref{sec:expsetup} describes the
experimental setup and baselines.
Section~\ref{sec:single-objective} presents
optimization results across single-objective,
multi-objective, and constrained settings,
including cross-model comparisons.
Section~\ref{sec:retrieval} isolates the effect
of retrieval tooling.
Section~\ref{sec:ibexablations} reports
robustness, sensitivity, and ablation studies,
and Section~\ref{sec:decision_process} analyzes
the agent's decision process through trajectory
and reasoning summaries.}

\subsection{\added{Experimental Setup and Baselines}}
\label{sec:expsetup}

Although the timeout per run depends on the circuit, each ``iteration'' is 
consistently defined as 
25 parallel runs, with the exception of the \texttt{JPEG} circuits, 
which are run with $12$ parallel runs each. Each of these parallel runs uses four 
vCPUs and 4 GB of RAM. These parallel runs write to different intermediate folders 
for logs and results, 
running Yosys and other steps in parallel without reusing intermediate steps. 
At the end of each iteration, a clean operation moves the logs and 
folders to a separate directory to avoid contamination.

Our experiments cover two settings. 
The first, ``4-variable'' setting involves optimizing over four 
parameters: {\bf Core Utilization}, {\bf TNS End Percent}, {\bf LB Add 
On Place Density}, and {\bf Clock Period}. The first three
are in the configuration files of the respective circuits; 
Clock Period is in the SDC file.
The second, ``12-variable'' setting has eight 
additional tunable parameters that guide the physical design flow 
more comprehensively. We distinguish three experimental cases.
\begin{itemize}
    \item \textbf{4-variable, no tool use}, where the LLM optimizes over four
    variables solely through prompt-tuning and in-context learning;

    \item  \textbf{4-variable with tool use}, where Bayesian optimization enhances
    performance; and

    \item \textbf{12-variable with tool use}, which incorporates
    all 12 parameters along with the Bayesian optimization tools.
\end{itemize}

Our experimental
choices reflect the LLM requiring more complex tools to tackle
higher-dimensional optimization problems. 
\added{All ORFS-agent experiments} involve either
$375$ (4-variable setting) or $600$ iterations
(12-variable setting)\added{; the OR-AutoTuner
baseline uses $375$ iterations in the
$4$-variable setting and $1000$ iterations in
the $12$-variable setting}.
(Table~\ref{tab:optimization_parameters} summarizes the tunable parameters and
their ranges.)

\begin{table}[!t]
\centering
\resizebox{1.0\columnwidth}{!}{%
\begin{tabular}{|l |l| c| c|}
\hline
\textbf{Parameter} & \textbf{Description} & \textbf{Type} & \textbf{Range} \\
\hline
\textbf{Clock Period}          & Target clock period (ns/ps)           & Float   & (0, $\infty$)        \\ \hline
\textbf{Core Utilization}      & \% core utilization                   & Integer & [20, 99]             \\ \hline
\textbf{TNS End Percent}       & \% violating endpoints to fix         & Integer & [0, 100]             \\ \hline
\textbf{Density Margin Add-On} & Global density margin increase        & Float   & [0.00, 0.99]         \\ \hline
\textbf{Global Padding}        & Global placement padding level        & Integer & [0, 3]               \\ \hline
\textbf{Detail Padding}        & Detailed placement padding level      & Integer & [0, 3]               \\ \hline
\textbf{Enable DPO}            & Detailed placement optimization       & Binary  & \{0, 1\}             \\ \hline
\textbf{Pin Layer Adjust}      & Routing adjust for metal2/3           & Float   & [0.2, 0.7]           \\ \hline
\textbf{Above Layer Adjust}    & Routing adjust for metal4 and above   & Float   & [0.2, 0.7]           \\ \hline
\textbf{Flatten Hierarchy}     & Flatten design hierarchy              & Binary  & \{0, 1\}             \\ \hline
\textbf{CTS Cluster Size}      & Number of sinks per CTS cluster       & Integer & [10, 40]             \\ \hline
\textbf{CTS Cluster Diameter}  & Physical span of each CTS cluster     & Integer & [80, 120]            \\
\hline
\end{tabular}
}
\caption{Overview of optimization parameters for both 4- and 12-parameter tuning.
The top four rows form the 4-parameter case.}
\label{tab:optimization_parameters}
\end{table}

The ORFS-agent is tasked with optimizing two specific figures of merit: the
post-route \textbf{effective clock period} ($ECP$) and \textbf{routed wirelength}
($WL$).\footnote{The unit of $ECP$ is $ns$ in SKY130HD and $ps$ in ASAP7; routed
wirelength is in $\mu m$; power is in $W$; and area is in $\mu m^2$.}
Due to potential timeouts during execution, these metrics may not always be
available. Therefore, ORFS-agent also records corresponding surrogates from the
clock tree synthesis (CTS) stage: \(ECP'\) and \(WL'\). Alongside
\(\{WL,ECP\}\), we track \textbf{instance area, instance count, total power}, and
a derived metric \textbf{PDP} (PDP = power \(\times\) \(ECP\)).
Throughout this work, we adopt the following notations:
\begin{itemize}
    \item $ECP^*$ (resp. $WL^*$) denotes the value of $ECP$ (resp. $WL$) when $WL$
    (resp. $ECP$) is optimized as the objective; and
    \item $ECP^{**}$ and $WL^{**}$ represent the values obtained when both metrics
    are jointly optimized in a multi-objective setting.
\end{itemize}

The ORFS-agent parameters are configured as follows:
\added{(i) unless otherwise stated, the LLM backend
is Claude Sonnet 4.6 \cite{anthropicClaude4}; we also
evaluate Moonshot Kimi K2.5 \cite{kimiK25} as an
open-weight model, and we retain the
Sonnet~3.5 results reported in~\cite{orfsagent_mlcad25} as a reference
operating point;}
(ii) temperature is \( 0.1 \);
(iii) nucleus and top-K sampling are disabled; and
(iv) to always output \( 25 \times 4/12 \) parameters (the number of 
parameters the LLM is optimizing over) in the specified format,
we utilize JSON Mode (Structured Outputs  
API).\footnote{In terms of API costs, the Sonnet~3.5 experiments reported in
\cite{orfsagent_mlcad25} (including failed runs) cost US \$48. Sonnet~3.5 has since been
overtaken by models which are simultaneously stronger and cheaper per token. As
noted above, approaches such as ChatEDA require fine-tuning, which costs
significantly more (by at least 10--20$\times$) and requires inference costs for
serving the custom model.}
\added{Cross-model and retrieval results are
presented in Sections~\ref{sec:single-objective}
and~\ref{sec:retrieval}.}

\label{sec:baseline}
\noindent
\added{\textbf{Baselines.}}
We first report baseline solution metrics obtained using the default ORFS flow
parameters at the cited commit hash. These baselines (Table~\ref{tab:benchmarks})
are used for normalization and for expressing relative improvements throughout
Section~\ref{sec:exp}. (For all metrics, \textbf{lower} is better.)
\added{For additional context on proprietary-tool QoR on ASAP7, see~\cite{jung2023ieee};
we do not attempt a direct comparison here due to differences in tool maturity and
resource budgets.}

\begin{table}[!t]
  \centering
  \resizebox{1.0\columnwidth}{!}{%
  \scriptsize
  \setlength{\tabcolsep}{3pt}
  \renewcommand{\arraystretch}{1.05}
  \begin{tabular}{@{}c c@{}}
  \begin{tabular}{|c|c|c|c|c|}
    \hline
    & CTSWL ($WL'$)  & CTSECP ($ECP'$) & $WL$ & $ECP$ \\ \hline
    SKY130HD-IBEX & 550963 & 10.84 & 808423 & 11.54 \\ \hline
    ASAP7-IBEX & 93005 & 1308 & 115285 & 1361 \\ \hline
    SKY130HD-AES & 428916 & 5.34 & 589825 & 4.72 \\ \hline
    ASAP7-AES & 61103 & 432 & 75438 & 460 \\ \hline
    SKY130HD-JPEG & 1199090 & 8.00 & 1374966 & 7.73 \\ \hline
    ASAP7-JPEG & 266510 & 1096 & 300326 & 1148 \\ \hline
  \end{tabular}
  &
  \begin{tabular}{|c|c|c|c|c|}
    \hline
    & Area & Count & Power & PDP \\ \hline
    SKY130HD-IBEX & 192784 & 20944 & 0.097 & 1.12 \\ \hline
    ASAP7-IBEX & 2729 & 21831 & 0.057 & 77.58 \\ \hline
    SKY130HD-AES & 122361 & 18324 & 0.411 & 1.94 \\ \hline
    ASAP7-AES & 2046 & 17693 & 0.149 & 68.54 \\ \hline
    SKY130HD-JPEG & 541327 & 65670 & 0.811 & 6.27 \\ \hline
    ASAP7-JPEG & 7904 & 68287 & 0.138 & 158.42 \\ \hline
  \end{tabular}
  \end{tabular}%
  }%
  \caption{Default ORFS baseline metrics (left)
  and derived quantities (right).}
  \label{tab:benchmarks}
\end{table}

\begin{table}[!t]
\centering
\resizebox{0.6\columnwidth}{!}{%
    \begin{tabular}{|c|c|cc|cc|cc|}
        \hline

\multirow{2}{*}{\textbf{Tech}} & \multirow{2}{*}{\textbf{Circuit}}
& \multicolumn{2}{c|}{\textbf{$WL$ Optimization}}
& \multicolumn{2}{c|}{\textbf{$ECP$ Optimization}}
& \multicolumn{2}{c|}{\textbf{Co-Optimization}} \\
\cline{3-8}
& & $WL$ & $ECP$* & $ECP$ & $WL$* & $WL$** & $ECP$** \\
\hline \hline

        \multicolumn{8}{|c|}{\textbf{OR-AutoTuner Performance: 375 Iterations, 4 Parameters}} \\ \hline
        \multirow{3}{*}{SKY130HD}
        & IBEX & 632918 & 14.52 & 10.17 & 811265 & 802786 & 10.49 \\
        & AES  & 514342 & 7.28  & 4.08 & 639876 & 564188 & 4.43 \\
        & JPEG & 1263740 & 8.25 & 6.77 & 1421110 & 1304829 & 7.24 \\
        \hline
        \multirow{3}{*}{ASAP7}
        & IBEX & 104622 & 1510 & 1252 & 109346 & 107821 & 1264 \\
        & AES  & 68943 & 1634 & 431 & 78152 & 72303 & 446 \\
        & JPEG & 269721 & 1172 & 950 & 284539 & 276453 & 978 \\
        \hline
       
        \hline  \hline

        \multicolumn{8}{|c|}{\textbf{OR-AutoTuner Performance: 1000 Iterations, 12 Parameters}} \\ \hline
        \multirow{3}{*}{SKY130HD}
        & IBEX & 630598 & 14.08 & 10.09 & 767451 & 750769 & 10.10 \\
        & AES  & 423730 & 5.02 & 3.81 & 563264 & 495781 & 3.89 \\
        & JPEG & 1070038 & 7.77 & 6.57 & 1295066 & 1089166 & 7.29 \\
        \hline
        \multirow{3}{*}{ASAP7}
        & IBEX & 96086 & 3244 & 1188 & 103081 & 100837 & 1192 \\
        & AES  & 66934 & 2220 & 430 & 77067 & 71086 & 443 \\
        & JPEG & 260683 & 1130 & 882 & 279621 & 273649 & 885 \\
        \hline
    \end{tabular}
    }%
    \caption{
        \textbf{OR-AutoTuner results.} 
        \textbf{Top block:} 375 iterations with 4 parameters. 
        \textbf{Bottom block:} 1000 iterations with 12 parameters. 
        Columns are grouped by optimization objective: $WL$-only ($WL$, $ECP$*),
        $ECP$-only ($ECP$, $WL$*), and co-optimization ($WL$**, $ECP$**), where
        $ECP$* denotes the achieved $ECP$ in a $WL$-only run, and $WL$* denotes the
        achieved $WL$ in an $ECP$-only run.
    }
    \label{tab:benchmarks_OR-AutoTuner}
\end{table}

\begin{table*}[!tp]
\centering
\added{\scriptsize
\setlength{\tabcolsep}{2.5pt}
\renewcommand{\arraystretch}{1.35}
\newcommand{\tri}[3]{%
  #1\,/\,{\color{son46color}#2}\,/\,{\color{kimicolor}#3}}
\resizebox{\textwidth}{!}{%
\begin{tabular}{|c|c|cc|cc|cc|}
\hline
\multirow{2}{*}{\textbf{Tech}}
& \multirow{2}{*}{\textbf{Circuit}}
& \multicolumn{2}{c|}{\textbf{$WL$ Optimization}}
& \multicolumn{2}{c|}{\textbf{$ECP$ Optimization}}
& \multicolumn{2}{c|}{\textbf{Co-Optimization}} \\
\cline{3-8}
& & $WL$ & $ECP$* & $ECP$ & $WL$*
& $WL$** & $ECP$** \\
\hline \hline
\multicolumn{8}{|c|}{\textbf{375 Iterations,
4 Parameters \emph{without} Tool Use}} \\ \hline
\multirow{3}{*}{SKY130HD}
& IBEX & \tri{732793}{627048}{627511}
& \tri{11.47}{11.40}{11.37}
& \tri{10.82}{10.06}{10.08}
& \tri{815984}{800800}{802315}
& \tri{792947}{786272}{784779}
& \tri{11.00}{10.42}{10.39} \\
& AES & \tri{531721}{507707}{508338}
& \tri{5.33}{5.27}{5.28}
& \tri{4.07}{4.02}{4.03}
& \tri{651219}{634252}{635194}
& \tri{637490}{559185}{561136}
& \tri{4.08}{4.03}{4.04} \\
& JPEG & \tri{1257231}{1244250}{1246184}
& \tri{7.28}{7.20}{7.20}
& \tri{6.69}{6.65}{6.63}
& \tri{1371607}{1353914}{1356490}
& \tri{1344380}{1287401}{1292341}
& \tri{6.77}{6.72}{6.71} \\
\hline
\multirow{3}{*}{ASAP7}
& IBEX & \tri{106044}{103679}{103724}
& \tri{1324}{1313}{1308}
& \tri{1281}{1238}{1236}
& \tri{113854}{108302}{108227}
& \tri{114576}{106446}{106806}
& \tri{1287}{1251}{1248} \\
& AES & \tri{70955}{68160}{68316}
& \tri{446}{442}{443}
& \tri{430}{427}{427}
& \tri{72427}{71842}{71967}
& \tri{71644}{70800}{70783}
& \tri{440}{436}{437} \\
& JPEG & \tri{277654}{266159}{266750}
& \tri{1084}{1075}{1077}
& \tri{1016}{939}{939}
& \tri{289031}{281321}{282107}
& \tri{286142}{273933}{273581}
& \tri{1069}{969}{968} \\
\hline \hline
\multicolumn{8}{|c|}{\textbf{375 Iterations,
4 Parameters \emph{with} Tool Use}} \\ \hline
\multirow{3}{*}{SKY130HD}
& IBEX & \tri{726874}{625919}{624754}
& \tri{10.91}{10.83}{10.81}
& \tri{10.74}{10.08}{10.11}
& \tri{752566}{747194}{745490}
& \tri{736275}{729355}{727958}
& \tri{10.84}{10.42}{10.39} \\
& AES & \tri{526782}{507707}{508756}
& \tri{5.12}{5.08}{5.06}
& \tri{4.02}{3.97}{3.97}
& \tri{601492}{594846}{594606}
& \tri{572966}{557811}{557041}
& \tri{4.06}{4.02}{4.03} \\
& JPEG & \tri{1268544}{1248088}{1252445}
& \tri{7.08}{7.00}{7.00}
& \tri{6.64}{6.58}{6.59}
& \tri{1327854}{1311247}{1313711}
& \tri{1302494}{1287974}{1292298}
& \tri{6.74}{6.66}{6.65} \\
\hline
\multirow{3}{*}{ASAP7}
& IBEX & \tri{101788}{100475}{100640}
& \tri{1298}{1287}{1286}
& \tri{1242}{1229}{1230}
& \tri{112788}{108166}{108337}
& \tri{106854}{105856}{105659}
& \tri{1254}{1245}{1243} \\
& AES & \tri{70288}{68305}{68301}
& \tri{448}{444}{443}
& \tri{435}{426}{426}
& \tri{72144}{71464}{71535}
& \tri{71422}{70603}{70798}
& \tri{442}{438}{437} \\
& JPEG & \tri{270184}{267603}{267312}
& \tri{968}{958}{962}
& \tri{952}{940}{938}
& \tri{283544}{280310}{280441}
& \tri{273126}{270304}{271346}
& \tri{958}{946}{947} \\
\hline \hline
\multicolumn{8}{|c|}{\textbf{600 Iterations,
12 Parameters \emph{with} Tool Use}} \\ \hline
\multirow{3}{*}{SKY130HD}
& IBEX & \tri{657091}{624941}{624783}
& \tri{12.25}{12.15}{12.13}
& \tri{10.42}{9.99}{10.02}
& \tri{721099}{713317}{713350}
& \tri{706087}{697308}{697716}
& \tri{11.06}{10.00}{10.01} \\
& AES & \tri{483426}{419113}{418071}
& \tri{4.82}{4.77}{4.77}
& \tri{3.96}{3.77}{3.78}
& \tri{539878}{534510}{534879}
& \tri{500296}{490621}{489291}
& \tri{4.06}{3.84}{3.84} \\
& JPEG & \tri{1237829}{1061138}{1062076}
& \tri{6.94}{6.87}{6.88}
& \tri{6.54}{6.47}{6.49}
& \tri{1308171}{1278360}{1279617}
& \tri{1254030}{1079030}{1081663}
& \tri{6.62}{6.55}{6.54} \\
\hline
\multirow{3}{*}{ASAP7}
& IBEX & \tri{97305}{95244}{95448}
& \tri{1278}{1261}{1263}
& \tri{1248}{1174}{1172}
& \tri{105562}{101929}{101752}
& \tri{102205}{99537}{99673}
& \tri{1260}{1181}{1178} \\
& AES & \tri{69824}{66448}{66355}
& \tri{456}{451}{452}
& \tri{432}{425}{426}
& \tri{74582}{73620}{73754}
& \tri{71084}{70397}{70235}
& \tri{438}{433}{433} \\
& JPEG & \tri{270963}{257931}{258774}
& \tri{905}{894}{896}
& \tri{878}{869}{870}
& \tri{275944}{272385}{272764}
& \tri{271822}{269410}{268789}
& \tri{882}{872}{873} \\
\hline
\end{tabular}%
}%
}%
\caption{\added{\textbf{ORFS-agent results (no retrieval)}
for various tuning settings across three backends.
Each cell shows
Sonnet~3.5\,/\,{\color{son46color}Sonnet~4.6}\,/\,%
{\color{kimicolor}Kimi~K2.5}.
\textbf{Top:} 375 iterations, 4 parameters (no tool);
\textbf{Middle:} 375 iterations, 4 parameters
(with tool);
\textbf{Bottom:} 600 iterations, 12 parameters
(with tool).
Column meanings follow
Table~\ref{tab:benchmarks_OR-AutoTuner}: $ECP$* is the
achieved $ECP$ in a $WL$-only run, $WL$* is the
achieved $WL$ in an $ECP$-only run, and
($WL$**, $ECP$**) are the co-optimization outcomes.}}
\label{tab:benchmarks_results}
\end{table*}

\begin{table*}[!t]
  \centering
  \scriptsize
  \setlength{\tabcolsep}{4pt}
  \renewcommand{\arraystretch}{1.05}
  \begin{tabular}{|c|c|c|c|c|c|c|}
    \hline
    \multirow{2}{*}{\textbf{Tech}} &
    \multirow{2}{*}{\textbf{Circuit}} &
    \multirow{2}{*}{\textbf{Objective}} &
    \multicolumn{4}{c|}{\textbf{Normalized objective
    (lower is better)}} \\
    \cline{4-7}
    & & &
    \textbf{OR-AutoTuner} &
    \textbf{Sonnet~3.5} &
    \textbf{Sonnet~4.6} &
    \textbf{Kimi~K2.5} \\
    \hline \hline
    \multirow{3}{*}{SKY130HD}
    & \multirow{3}{*}{IBEX}
      & $WL$-opt (\(WL/WL_\alpha\))
      & 0.780 & 0.813
      & \textbf{0.773} & \textbf{0.773} \\
    & & $ECP$-opt (\(ECP/ECP_\alpha\))
      & 0.874 & 0.903
      & \textbf{0.866} & 0.868 \\
    & & Co-opt
      (\(0.5(WL/WL_\alpha + ECP/ECP_\alpha)\))
      & 0.902 & 0.916
      & \textbf{0.865} & \textbf{0.865} \\
    \hline
    \multirow{3}{*}{SKY130HD}
    & \multirow{3}{*}{AES}
      & $WL$-opt (\(WL/WL_\alpha\))
      & 0.718 & 0.820
      & 0.711 & \textbf{0.709} \\
    & & $ECP$-opt (\(ECP/ECP_\alpha\))
      & 0.807 & 0.839
      & \textbf{0.799} & 0.801 \\
    & & Co-opt
      (\(0.5(WL/WL_\alpha + ECP/ECP_\alpha)\))
      & 0.832 & 0.854
      & 0.823 & \textbf{0.822} \\
    \hline
    \multirow{3}{*}{SKY130HD}
    & \multirow{3}{*}{JPEG}
      & $WL$-opt (\(WL/WL_\alpha\))
      & 0.778 & 0.900
      & \textbf{0.772} & \textbf{0.772} \\
    & & $ECP$-opt (\(ECP/ECP_\alpha\))
      & 0.850 & 0.846
      & \textbf{0.837} & 0.840 \\
    & & Co-opt
      (\(0.5(WL/WL_\alpha + ECP/ECP_\alpha)\))
      & 0.868 & 0.884
      & \textbf{0.816} & \textbf{0.816} \\
    \hline
    \multirow{3}{*}{ASAP7}
    & \multirow{3}{*}{IBEX}
      & $WL$-opt (\(WL/WL_\alpha\))
      & 0.833 & 0.844
      & \textbf{0.826} & 0.828 \\
    & & $ECP$-opt (\(ECP/ECP_\alpha\))
      & 0.873 & 0.917
      & 0.863 & \textbf{0.861} \\
    & & Co-opt
      (\(0.5(WL/WL_\alpha + ECP/ECP_\alpha)\))
      & 0.875 & 0.906
      & 0.866 & \textbf{0.865} \\
    \hline
    \multirow{3}{*}{ASAP7}
    & \multirow{3}{*}{AES}
      & $WL$-opt (\(WL/WL_\alpha\))
      & 0.887 & 0.926
      & 0.881 & \textbf{0.880} \\
    & & $ECP$-opt (\(ECP/ECP_\alpha\))
      & 0.935 & 0.939
      & \textbf{0.924} & 0.926 \\
    & & Co-opt
      (\(0.5(WL/WL_\alpha + ECP/ECP_\alpha)\))
      & 0.953 & 0.947
      & 0.937 & \textbf{0.936} \\
    \hline
    \multirow{3}{*}{ASAP7}
    & \multirow{3}{*}{JPEG}
      & $WL$-opt (\(WL/WL_\alpha\))
      & 0.868 & 0.902
      & \textbf{0.859} & 0.862 \\
    & & $ECP$-opt (\(ECP/ECP_\alpha\))
      & 0.768 & 0.765
      & \textbf{0.757} & 0.758 \\
    & & Co-opt
      (\(0.5(WL/WL_\alpha + ECP/ECP_\alpha)\))
      & 0.841 & 0.837
      & \textbf{0.828} & \textbf{0.828} \\
    \hline \hline
    \multicolumn{2}{|c|}{\textbf{Geometric mean}}
    & $WL$-opt (\(WL/WL_\alpha\))
    & 0.809 & 0.866
    & \textbf{0.801} & 0.802 \\
    \hline
    \multicolumn{2}{|c|}{\textbf{Geometric mean}}
    & $ECP$-opt (\(ECP/ECP_\alpha\))
    & 0.850 & 0.866
    & \textbf{0.839} & 0.841 \\
    \hline
    \multicolumn{2}{|c|}{\textbf{Geometric mean}}
    & Co-opt
    (\(0.5(WL/WL_\alpha + ECP/ECP_\alpha)\))
    & 0.878 & 0.890
    & 0.855 & \textbf{0.854} \\
    \hline
  \end{tabular}
  \caption{\added{\textbf{Cross-model comparison of
  single- and multi-objective QoR.}
  ORFS-agent rows use 600 iterations and 12 parameters
  with tool use (no retrieval); OR-AutoTuner uses 1000
  iterations and 12 parameters.
  Each entry is normalized by the same default ORFS
  baseline for that circuit in Table~\ref{tab:benchmarks}
  (lower is better), so the three ORFS-agent backend
  columns can be compared directly as
  Sonnet~3.5$\rightarrow$Sonnet~4.6$\rightarrow$Kimi~K2.5.
  For co-optimization, we report
  \(0.5(WL/WL_\alpha + ECP/ECP_\alpha)\). Bold
  indicates the best value per row.}}
  \label{tab:thinking_model_comparison}
\end{table*}

\begin{figure*}[!t]
    \centering
    \includegraphics[width=0.96\textwidth]{%
    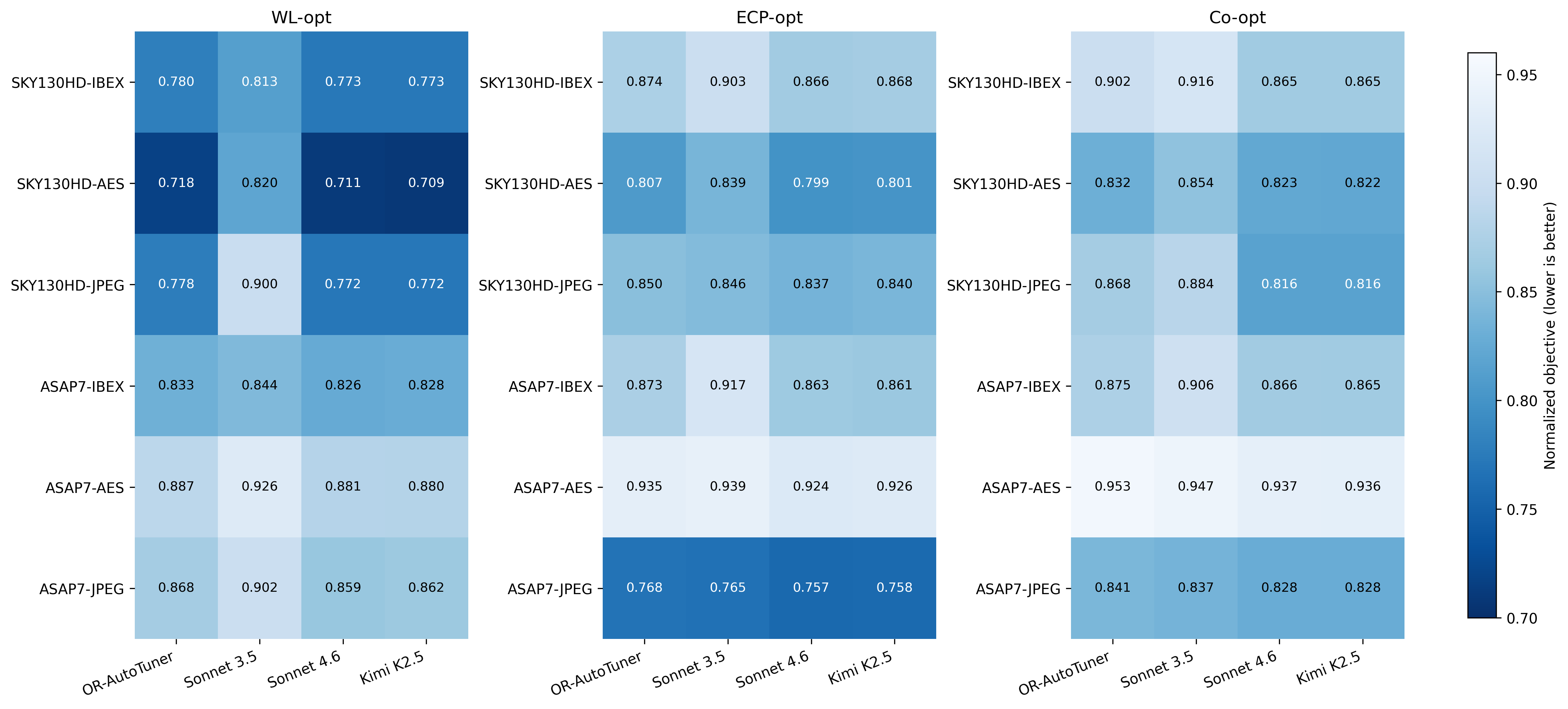}
    \caption{\added{Heatmap summary of
    Table~\ref{tab:thinking_model_comparison}.
    Each panel shows the normalized objective for
    all six circuits under OR-AutoTuner,
    Sonnet~3.5, Sonnet~4.6, and Kimi~K2.5 at the
    \(600\)-iteration, \(12\)-parameter setting.
    Lower values are better.}}
    \Description{Three heatmaps summarizing
    normalized objectives across six circuits and
    four methods.}
    \label{fig:cross_model_heatmaps}
\end{figure*}

\subsection{\added{Optimization Results}}
\label{sec:single-objective}
\label{subsec:multi-objective}
\label{sec:addnl}

In the single-objective optimization scenario 
(e.g., minimizing routed wirelength \( WL \)), 
the LLM agent receives an additional natural language instruction appended 
to its prompt. Specifically, we append the following instruction:

\begin{quote}
\small
\textcolor{instructionviolet}{ \em There is only a single objective
to optimize in this problem. You observe that 
the value inside the final JSON is $WL$, while the corresponding 
baseline value is $WL_\alpha$. Your effective loss to minimize 
should be $\frac{WL}{WL_\alpha}$.}
\end{quote}

\noindent
where $WL_\alpha$ denotes the routed wirelength obtained using the 
default configuration of OpenROAD-flow-scripts and serves as a fixed baseline for 
normalization.\footnote{The baseline results obtained using the default 
configuration of OpenROAD-flow-scripts are shown in Section~\ref{sec:baseline}.} 

However, due to runtime timeouts, 
typically occurring during the detailed routing stage,
the routed wirelength $WL$ may not always be available. 
In contrast, the estimated wirelength at the end of CTS  $WL'$
almost always exists. If $WL$ is absent but $WL'$
exists, we include the following instruction:
\begin{quote}
\small
\textcolor{instructionviolet}{
\em You do not observe the correct value $WL$,
but you observe a strong surrogate $WL'$, which
you should use as a signal. Your effective loss
should be $\frac{WL'}{WL'_\alpha}$.}
\end{quote}

\noindent
where $WL'_\alpha$ denotes post-CTS wirelength obtained under 
the same default configuration. 
This surrogate-based guidance enables the LLM to continue learning from 
partial information while maintaining alignment with the optimization objective.

The experimental results are shown in Table~\ref{tab:benchmarks_results}.
\added{Table~\ref{tab:benchmarks_results} reports the Sonnet~3.5 reference runs
from~\cite{orfsagent_mlcad25}. Table~\ref{tab:thinking_model_comparison} is the
primary side-by-side comparison across Sonnet~3.5, Sonnet~4.6, and Kimi~K2.5
under the same 600-iteration, 12-parameter, tool-enabled setting. Because each
row uses one fixed circuit-specific normalization baseline, reading the three
ORFS-agent backend columns isolates the backend change itself rather than a
change in scaling. In that table, the geometric-mean normalized \(WL\)
objective decreases from \(0.866\) with Sonnet~3.5 to \(0.801/0.802\), and the
geometric-mean normalized \(ECP\) objective decreases from \(0.866\) to
\(0.839/0.841\). Figure~\ref{fig:cross_model_heatmaps} visualizes the same
comparison.}

\noindent
With respect to Table~\ref{tab:benchmarks_results},
key observations are as follows.
\begin{itemize}
    \item \textbf{Comparison between default ORFS and
    ORFS-agent}: ORFS-agent (600 iterations,
    12 parameters with tool use) achieves an average
    improvement of \added{$13.3\%$ in wirelength and
    $13.2\%$ in $ECP$} when configured for the
    respective optimization objective.
    \item \textbf{Comparison between OR-AutoTuner and
    ORFS-agent}: Despite using 40\% fewer iterations,
    ORFS-agent (600 iterations, 12 parameters, with tool
    use) improves $12$ of the $24$ single-objective
    endpoint metrics relative to OR-AutoTuner.

	    \item \textbf{Impact of tool use} (375 iterations
	    and 4 parameters): ORFS-agent with tool use
	    dominates ORFS-agent without tool use in 5 out of
	    6 cases, for both wirelength and effective clock
	    period optimization tasks.

    \item We notice that ORFS-agent does \textbf{not}
    significantly worsen the metric not being optimized.
    For instance, when optimizing wirelength on
    ASAP7-IBEX, the $ECP$ obtained by ORFS-agent
    (600 iterations, 12 parameters with tool use) is
    $40\%$ of that obtained by OR-AutoTuner.
	    We hypothesize this is from the $ECP$ values
	    entering the agent's context window, causing it to
	    avoid actions that would worsen $ECP$ at the same time.

    \item \added{\textbf{Across model backends
    (Table~\ref{tab:thinking_model_comparison}).}
    Sonnet~4.6 and Kimi~K2.5 both outperform
    OR-AutoTuner on geometric mean while using
    \(40\%\) fewer iterations: Sonnet~4.6 improves
    the normalized \(WL\), \(ECP\), and
    co-optimization objectives by \(1.0\%\),
    \(1.3\%\), and \(2.6\%\), respectively, while
    Kimi~K2.5 improves them by \(0.9\%\),
    \(1.1\%\), and \(2.7\%\). Relative to
    Sonnet~3.5, the same geometric-mean objectives
    fall from \(0.866\) to \(0.801/0.802\) for
    \(WL\), from \(0.866\) to \(0.839/0.841\) for
    \(ECP\), and from \(0.890\) to \(0.855/0.854\)
    for co-optimization. The open-weight
	    Kimi~K2.5 remains within \(0.24\%\) of
	    Sonnet~4.6 on all three geometric-mean
	    objectives and slightly improves the
	    co-optimization average. Figure~\ref{fig:cross_model_heatmaps}
	    visualizes these backend comparisons.}
\end{itemize}

\begin{addedblock}
\begin{figure*}[t]
    \centering
    \includegraphics[width=0.92\textwidth]{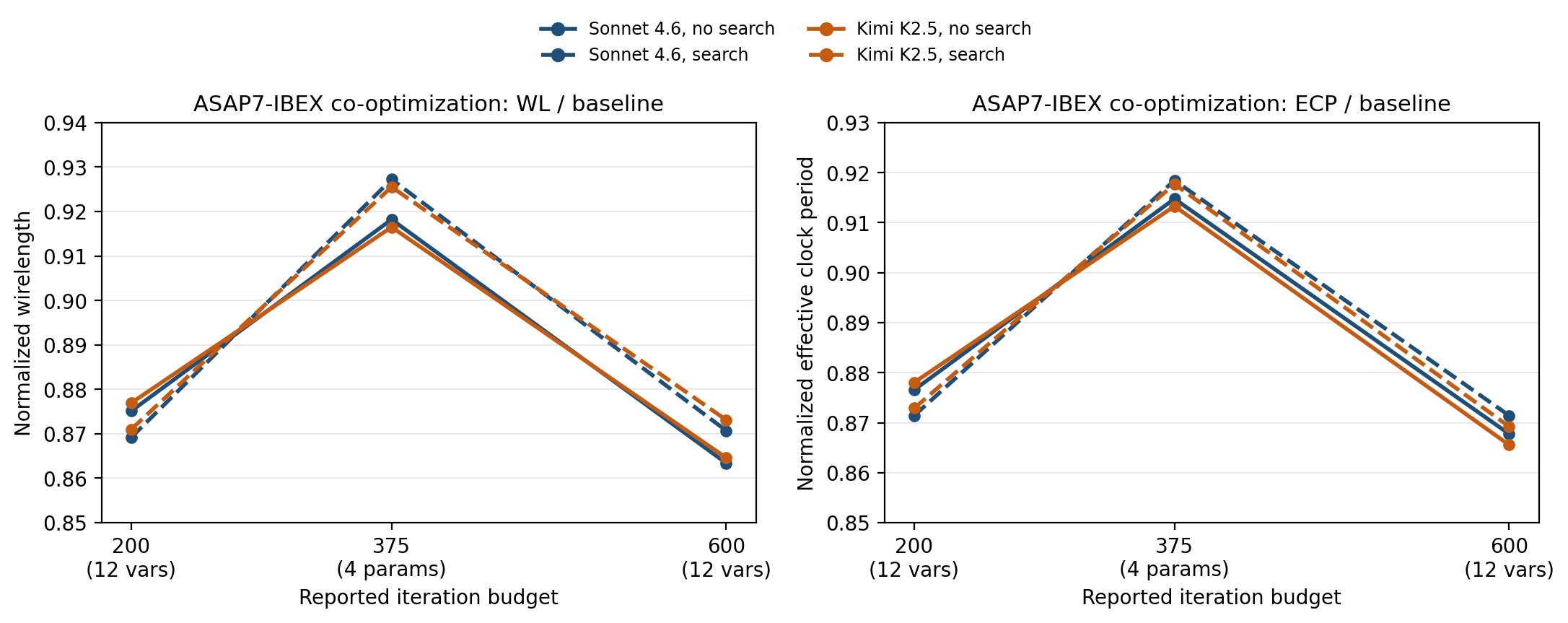}
    \caption{Thinking-model ASAP7-IBEX co-optimization checkpoints at key evaluation
    budgets. Each point is copied directly from the reported Sonnet 4.6 and
    Kimi K2.5 tables: Table~\ref{tab:benchmarks_results_lowiter} provides the
    $200$-iteration checkpoints (12 parameters), while
    Table~\ref{tab:benchmarks_results_merged} provides the $375$-iteration
    checkpoints (4 parameters, tool use) and $600$-iteration checkpoints
    (12 parameters, tool use), both without and with retrieval. Wirelength and effective clock period are
    normalized by the default ASAP7-IBEX baseline in Table~\ref{tab:benchmarks}
    (\(WL_\alpha = 115285\), \(ECP_\alpha = 1361\)); lower is better.}
    \Description{Two line charts showing normalized ASAP7-IBEX co-optimization checkpoints for Sonnet 4.6 and Kimi K2.5, with and without search, at reported budgets of 200, 375, and 600 iterations relative to the default baseline.}
    \label{fig:3x3_collage}
\end{figure*}
\end{addedblock}

\noindent
\added{\textbf{Multi-objective formulation.}}
In the case of multiple losses -- e.g., $WL$ and
$ECP$ -- we instruct the LLM to consider the loss:
\begin{equation}
    \frac{WL}{WL_\alpha} + \frac{ECP}{ECP_\alpha}
\end{equation}
with the same substitution, if applicable, for $ECP'$, $WL'$ if
any of these are unavailable, with the same phrasing adopted above.
The experimental results are shown in Table~\ref{tab:benchmarks_results}.
\added{The same co-optimization task is compared across Sonnet~3.5,
Sonnet~4.6, Kimi~K2.5, and OR-AutoTuner in
Table~\ref{tab:thinking_model_comparison}. On geometric mean, the normalized
co-optimization objective improves from \(0.890\) with Sonnet~3.5 to
\(0.855\) with Sonnet~4.6 and \(0.854\) with Kimi~K2.5, compared with
\(0.878\) for OR-AutoTuner.}
Key results are as follows.
\begin{itemize}
    \item \textbf{Sustained gains.} Relative to the
    default ORFS baseline, the Sonnet~3.5 reference
    backend improves simultaneous \(ECP\) and \(WL\) by
    roughly \(11.3\%\) and \(10.5\%\), respectively, in
    the \(600\)-iteration, \(12\)-parameter,
    tool-enabled setting.

    \item \textbf{No worsening.} As with
    single-objective optimization, ORFS-agent does not
    worsen any metric -- either $ECP$ or wirelength,
    relative to baselines. In fact, both QoR metrics
	    improve \textbf{together.}

    \item \added{\textbf{Comparison with OR-AutoTuner
    and thinking models.} In 6 of the 12 circuit-level
    co-optimization comparisons, ORFS-agent outperforms
    OR-AutoTuner despite using fewer iterations.
    Table~\ref{tab:thinking_model_comparison} then shows
    that Sonnet~4.6 and Kimi~K2.5 improve the
    geometric-mean co-optimization objective further, to
    \(0.855\) and \(0.854\), respectively.}

\end{itemize}

\begin{figure}[!ht]
    \centering
    \includegraphics[width=\columnwidth]{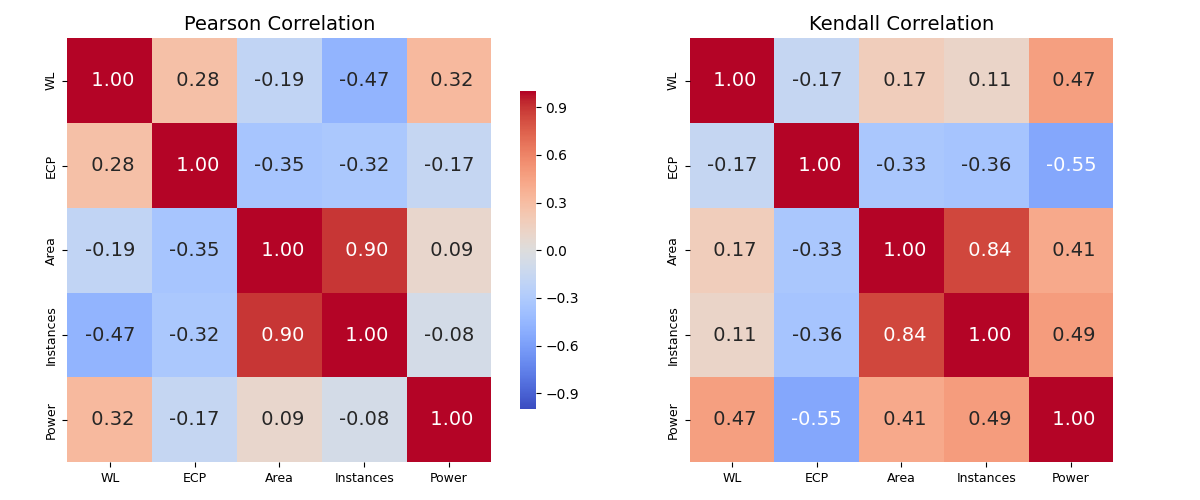}
    \caption{Correlation of metrics for ASAP7-IBEX: 
    wirelength ($WL$), 
    effective clock period ($ECP$),
    instance area (Area),
    instance count (Instances) and power (Power). 
    Pearson correlation captures linear 
    relationships; Kendall correlation captures rank-based relationships.}
    \Description{Correlation heatmap for ASAP7-IBEX comparing Pearson and Kendall correlations among wirelength, effective clock period, area, instance count, and power.}
    
    \label{fig:corrheatmap}
\end{figure}

We visualize the thinking-model ASAP7-IBEX co-optimization checkpoints in
Figure~\ref{fig:3x3_collage}. The figure focuses on
$WL$ and $ECP$, which are reported consistently across the Sonnet 4.6 and Kimi
K2.5 runs. Two trends stand out: both models remain below the default ORFS
baseline at every reported checkpoint, and search helps most at $200$
iterations, whereas the strongest $600$-iteration endpoints are obtained
without search.
Table~\ref{tab:latest_model_trajectory_abs} restates the reported
\(15\)- and \(30\)-iteration thinking-model checkpoints in the same
\(5\) to \(30\) serial-iteration format as the legacy ASAP7-IBEX trajectory
tables, and
Tables~\ref{tab:reasoning_sonnet46}--\ref{tab:reasoning_search_200} summarize
the aligned reasoning patterns. To better understand gains on auxiliary metrics (metrics that do not
directly enter the optimization function), we examine correlational properties
of metrics in Figure~\ref{fig:corrheatmap}.
We look at how other metrics vary in an optimization run as we carry out ORFS-level
runs for ASAP7-IBEX under different parameters and differing optimization goals -- 
namely wirelength, $ECP$, or a combination of the two as defined above. 
Some metrics do not ``come for free" -- $ECP$ and power, for instance, have an inverse relationship with each other.

\label{subsec:constopt}
\noindent
\added{\textbf{Constrained optimization.}}
A human engineer often has goals that differ
from the utility functions used by
OR-AutoTuner. For example, a human might be
instructed: ``Optimize metric \( X \) as much as
possible, while not allowing metrics
\( A, B, C \) to deteriorate by more than
\( 2\% \).'' This percentage value is with
reference to the baseline values in
Table~\ref{tab:benchmarks}.
This constrained optimization is not
straightforward with existing OR-AutoTuner
setups. But by contrast, LLMs can attempt such optimization by
incorporating these constraints directly into
their context.
\added{Table~\ref{tab:constrainedoptORFSAgent-mainpaper}
compares ORFS-agent (Sonnet~3.5 and Kimi~K2.5)
against OR-AutoTuner under \( 2\% \) and
\( 4\% \) leeway constraints on Area, Instance
Count, Power, and PDP.} We term the maximum
worsening allowed as ``leeway.''
\added{ORFS-agent respects the constraints while
simultaneously improving multiple metrics.
Notably, OR-AutoTuner violates the leeway
bounds more frequently than ORFS-agent,
particularly for $ECP$ under $WL$ optimization
in the 12-parameter setting.}

\begin{table*}[!ht]
\centering
\added{\scriptsize
\begin{tabular}{|c|c|c|c|c|}
\hline
\multicolumn{5}{|c|}{\textbf{ORFS-agent: 4 params,
no tools / OR-AutoTuner: 375 iter, 4 params}}
\\ \hline
\multirow{2}{*}{Metric}
& \multicolumn{2}{c|}{$WL$ Optimization}
& \multicolumn{2}{c|}{$ECP$ Optimization}
\\ \cline{2-5}
& 2\% leeway & 4\% leeway
& 2\% leeway & 4\% leeway \\ \hline
$WL$
& {\color{oratcolor}-9.03}\,/\,{-8.18}\,/\,{-8.94}
& {\color{oratcolor}-5.69}\,/\,{-8.51}\,/\,{-9.28}
& {\color{oratcolor}-3.23}\,/\,{-7.62}\,/\,{-8.10}
& {\color{oratcolor}-1.41}\,/\,{-6.34}\,/\,{-6.88}
\\ \hline
$ECP$
& \textcolor{red}{{\color{oratcolor}1.40}}\,/\,{-1.07}\,/\,{-1.34}
& \textcolor{red}{{\color{oratcolor}1.03}}\,/\,\textcolor{red}{2.51}\,/\,\textcolor{red}{1.96}
& {\color{oratcolor}-1.31}\,/\,{-2.98}\,/\,{-3.34}
& {\color{oratcolor}-2.60}\,/\,{-3.06}\,/\,{-3.41}
\\ \hline
Area
& {\color{oratcolor}-1.59}\,/\,{-1.36}\,/\,{-1.49}
& {\color{oratcolor}-1.22}\,/\,{-1.68}\,/\,{-1.81}
& {\color{oratcolor}-0.39}\,/\,{-1.18}\,/\,{-1.27}
& {\color{oratcolor}-0.08}\,/\,{-0.80}\,/\,{-0.92}
\\ \hline
Count
& {\color{oratcolor}-1.80}\,/\,{-1.55}\,/\,{-1.63}
& {\color{oratcolor}-1.71}\,/\,{-1.70}\,/\,{-1.82}
& \textcolor{red}{{\color{oratcolor}0.58}}\,/\,{-1.35}\,/\,{-1.43}
& {\color{oratcolor}-0.02}\,/\,{-1.17}\,/\,{-1.25}
\\ \hline
PDP
& {\color{oratcolor}-4.16}\,/\,{-2.03}\,/\,{-2.46}
& {\color{oratcolor}-7.87}\,/\,{-9.04}\,/\,{-9.72}
& {\color{oratcolor}-0.55}\,/\,{-5.12}\,/\,{-5.58}
& {\color{oratcolor}-0.06}\,/\,{-0.05}\,/\,{-0.32}
\\ \hline
Power
& {\color{oratcolor}-5.49}\,/\,{-0.97}\,/\,{-1.11}
& {\color{oratcolor}-8.81}\,/\,{-11.27}\,/\,{-10.83}
& \textcolor{red}{{\color{oratcolor}0.77}}\,/\,{-2.20}\,/\,{-2.46}
& \textcolor{red}{{\color{oratcolor}2.61}}\,/\,\textcolor{red}{3.11}\,/\,\textcolor{red}{2.47}
\\ \hline
\multicolumn{5}{|c|}{\textbf{ORFS-agent: 12 params,
tools / OR-AutoTuner: 1000 iter, 12 params}}
\\ \hline
\multirow{2}{*}{Metric}
& \multicolumn{2}{c|}{$WL$ Optimization}
& \multicolumn{2}{c|}{$ECP$ Optimization}
\\ \cline{2-5}
& 2\% leeway & 4\% leeway
& 2\% leeway & 4\% leeway \\ \hline
$WL$
& {\color{oratcolor}-12.53}\,/\,{-11.24}\,/\,{-11.78}
& {\color{oratcolor}-9.26}\,/\,{-12.05}\,/\,{-12.63}
& \textcolor{red}{{\color{oratcolor}1.59}}\,/\,{-7.9}\,/\,{-8.34}
& {\color{oratcolor}-7.04}\,/\,{-8.3}\,/\,{-8.79}
\\ \hline
$ECP$
& \textcolor{red}{{\color{oratcolor}1.07}}\,/\,{-3.02}\,/\,{-3.31}
& \textcolor{red}{{\color{oratcolor}2.46}}\,/\,{-2.12}\,/\,{-2.38}
& {\color{oratcolor}-0.46}\,/\,{-4.18}\,/\,{-4.54}
& {\color{oratcolor}-2.67}\,/\,{-4.46}\,/\,{-4.82}
\\ \hline
Area
& {\color{oratcolor}-1.86}\,/\,{-2.17}\,/\,{-2.26}
& {\color{oratcolor}-1.65}\,/\,{-1.92}\,/\,{-2.05}
& {\color{oratcolor}-0.49}\,/\,{-1.12}\,/\,{-1.24}
& {\color{oratcolor}-1.95}\,/\,\textcolor{red}{0.56}\,/\,\textcolor{red}{0.43}
\\ \hline
Count
& {\color{oratcolor}-2.0}\,/\,{-1.62}\,/\,{-1.71}
& {\color{oratcolor}-1.47}\,/\,{-1.74}\,/\,{-1.86}
& \textcolor{red}{{\color{oratcolor}0.16}}\,/\,{-1.31}\,/\,{-1.41}
& {\color{oratcolor}-1.57}\,/\,{-1.06}\,/\,{-1.15}
\\ \hline
PDP
& {\color{oratcolor}-7.17}\,/\,{-3.25}\,/\,{-3.62}
& {\color{oratcolor}-4.17}\,/\,{-0.27}\,/\,{-0.41}
& {\color{oratcolor}-1.10}\,/\,{-6.88}\,/\,{-7.42}
& {\color{oratcolor}-2.31}\,/\,{-0.77}\,/\,{-0.89}
\\ \hline
Power
& {\color{oratcolor}-8.15}\,/\,{-0.24}\,/\,{-0.36}
& {\color{oratcolor}-6.47}\,/\,\textcolor{red}{1.87}\,/\,\textcolor{red}{1.39}
& {\color{oratcolor}-0.65}\,/\,{-2.81}\,/\,{-3.07}
& \textcolor{red}{{\color{oratcolor}0.37}}\,/\,\textcolor{red}{3.86}\,/\,\textcolor{red}{2.94}
\\ \hline
\end{tabular}
}
\caption{\added{\textbf{Constrained optimization results (\% change).}
Cells show {\color{oratcolor}OR-AutoTuner}\,/\,Sonnet~3.5\,/\,%
{\color{kimicolor}Kimi~K2.5}. \textbf{Top:} ORFS-agent with 4 parameters and
no tools, compared against OR-AutoTuner at 375 iterations. \textbf{Bottom:}
ORFS-agent with 12 parameters and tools, compared against OR-AutoTuner at
1000 iterations. \textcolor{red}{Red} indicates worsening; ``leeway'' is the
maximum allowed worsening.}}
\label{tab:constrainedoptORFSAgent-mainpaper}
\label{tab:constrainedoptORAutoTuner}
\label{tab:constrainedoptORAutoTuner-12param}
\end{table*}

\begin{table*}[!t]
    \centering
    \scriptsize
    \setlength{\tabcolsep}{3pt}
    \renewcommand{\arraystretch}{1.03}

    \textbf{Retrieval enabled}\\[0.2em]
    \begin{minipage}[t]{0.49\textwidth}
    \centering
    \textbf{Sonnet 4.6}\\[0.2em]
    \resizebox{\linewidth}{!}{%
    \begin{tabular}{|c|c|cc|cc|cc|}
        \hline
        \multirow{2}{*}{\textbf{Tech}} & \multirow{2}{*}{\textbf{Circuit}}
        & \multicolumn{2}{c|}{\textbf{$WL$ Optimization}}
        & \multicolumn{2}{c|}{\textbf{$ECP$ Optimization}}
        & \multicolumn{2}{c|}{\textbf{Co-Optimization}} \\
        \cline{3-8}
        & & $WL$ & $ECP$* & $ECP$ & $WL$* & $WL$** & $ECP$** \\
        \hline \hline
        \multicolumn{8}{|c|}{\textbf{375 Iterations, 4 Parameters \emph{without} Tool Use}} \\ \hline
        \multirow{3}{*}{SKY130HD}
        & IBEX & 625998 & 11.44 & 10.13 & 801125 & 793987 & 10.40 \\
        & AES & 507913 & 5.27 & 4.04 & 641358 & 563599 & 4.03 \\
        & JPEG & 1251559 & 7.23 & 6.68 & 1358651 & 1288817 & 6.79 \\
        \hline
        \multirow{3}{*}{ASAP7}
        & IBEX & 104494 & 1310 & 1244 & 109312 & 107497 & 1256 \\
        & AES & 68931 & 447 & 426 & 71904 & 71566 & 439 \\
        & JPEG & 268319 & 1074 & 942 & 280992 & 275710 & 975 \\
        \hline
        \hline \hline
        \multicolumn{8}{|c|}{\textbf{375 Iterations, 4 Parameters \emph{with} Tool Use}} \\ \hline
        \multirow{3}{*}{SKY130HD}
        & IBEX & 624869 & 10.87 & 10.15 & 746497 & 736518 & 10.40 \\
        & AES & 507913 & 5.08 & 3.98 & 601525 & 562226 & 4.02 \\
        & JPEG & 1255435 & 7.03 & 6.62 & 1317626 & 1289388 & 6.72 \\
        \hline
        \multirow{3}{*}{ASAP7}
        & IBEX & 101131 & 1284 & 1236 & 109177 & 106897 & 1250 \\
        & AES & 69076 & 449 & 426 & 71526 & 71366 & 440 \\
        & JPEG & 269763 & 957 & 942 & 279982 & 272060 & 950 \\
        \hline
        \hline \hline
        \multicolumn{8}{|c|}{\textbf{600 Iterations, 12 Parameters \emph{with} Tool Use}} \\ \hline
        \multirow{3}{*}{SKY130HD}
        & IBEX & 623895 & 12.20 & 10.07 & 712650 & 704178 & 9.98 \\
        & AES & 418574 & 4.77 & 3.79 & 540505 & 494500 & 3.84 \\
        & JPEG & 1067360 & 6.90 & 6.50 & 1282767 & 1080212 & 6.61 \\
        \hline
        \multirow{3}{*}{ASAP7}
        & IBEX & 95993 & 1263 & 1180 & 102882 & 100378 & 1186 \\
        & AES & 67196 & 456 & 425 & 73600 & 71157 & 436 \\
        & JPEG & 260019 & 893 & 871 & 272495 & 271157 & 876 \\
        \hline
        \hline
    \end{tabular}%
    }
    \end{minipage}
    \hfill
    \begin{minipage}[t]{0.49\textwidth}
    \centering
    \textbf{Kimi K2.5}\\[0.2em]
    \resizebox{\linewidth}{!}{%
    \begin{tabular}{|c|c|cc|cc|cc|}
        \hline
        \multirow{2}{*}{\textbf{Tech}} & \multirow{2}{*}{\textbf{Circuit}}
        & \multicolumn{2}{c|}{\textbf{$WL$ Optimization}}
        & \multicolumn{2}{c|}{\textbf{$ECP$ Optimization}}
        & \multicolumn{2}{c|}{\textbf{Co-Optimization}} \\
        \cline{3-8}
        & & $WL$ & $ECP$* & $ECP$ & $WL$* & $WL$** & $ECP$** \\
        \hline \hline
        \multicolumn{8}{|c|}{\textbf{375 Iterations, 4 Parameters \emph{without} Tool Use}} \\ \hline
        \multirow{3}{*}{SKY130HD}
        & IBEX & 626461 & 11.42 & 10.15 & 801565 & 792493 & 10.37 \\
        & AES & 507684 & 5.29 & 4.05 & 642299 & 565551 & 4.04 \\
        & JPEG & 1253493 & 7.23 & 6.66 & 1363080 & 1293757 & 6.78 \\
        \hline
        \multirow{3}{*}{ASAP7}
        & IBEX & 104540 & 1308 & 1242 & 109238 & 107857 & 1253 \\
        & AES & 69087 & 448 & 426 & 72029 & 71549 & 440 \\
        & JPEG & 268910 & 1075 & 941 & 281778 & 275358 & 973 \\
        \hline
        \hline \hline
        \multicolumn{8}{|c|}{\textbf{375 Iterations, 4 Parameters \emph{with} Tool Use}} \\ \hline
        \multirow{3}{*}{SKY130HD}
        & IBEX & 625007 & 10.85 & 10.18 & 744793 & 735121 & 10.37 \\
        & AES & 508101 & 5.07 & 3.99 & 601285 & 561455 & 4.03 \\
        & JPEG & 1259792 & 7.03 & 6.62 & 1320090 & 1293712 & 6.72 \\
        \hline
        \multirow{3}{*}{ASAP7}
        & IBEX & 101434 & 1283 & 1237 & 109348 & 106700 & 1249 \\
        & AES & 69072 & 448 & 426 & 71597 & 71562 & 439 \\
        & JPEG & 269472 & 961 & 941 & 280114 & 273102 & 952 \\
        \hline
        \hline \hline
        \multicolumn{8}{|c|}{\textbf{600 Iterations, 12 Parameters \emph{with} Tool Use}} \\ \hline
        \multirow{3}{*}{SKY130HD}
        & IBEX & 623737 & 12.18 & 10.09 & 712683 & 704586 & 10.00 \\
        & AES & 418434 & 4.77 & 3.80 & 540874 & 493170 & 3.84 \\
        & JPEG & 1068297 & 6.91 & 6.52 & 1285838 & 1082845 & 6.60 \\
        \hline
        \multirow{3}{*}{ASAP7}
        & IBEX & 96197 & 1261 & 1178 & 102583 & 100656 & 1183 \\
        & AES & 67103 & 457 & 425 & 73817 & 70995 & 436 \\
        & JPEG & 260862 & 895 & 872 & 272445 & 270536 & 878 \\
        \hline
        \hline
    \end{tabular}%
    }
    \end{minipage}

    \caption{\added{\textbf{ORFS-agent results with
    retrieval enabled} under \textbf{Sonnet~4.6} and
    \textbf{Kimi~K2.5}. Tranches correspond to:
    \textbf{(Top)} 375 iterations, 4 parameters,
    no tool use; \textbf{(Middle)} 375 iterations,
    4 parameters, with tool use; and
    \textbf{(Bottom)} 600 iterations, 12 parameters,
    with tool use. No-retrieval results are in
    Table~\ref{tab:benchmarks_results}. Column
    meanings follow
    Table~\ref{tab:benchmarks_OR-AutoTuner}.}}
    \label{tab:benchmarks_results_merged}
\end{table*}

\subsection{\added{Effect of Retrieval}}
\label{sec:retrieval}

\added{To test whether external knowledge aids
optimization, we enabled two Retrieval tools
-- web search (Brave) and scholarly lookup (OpenAlex)
-- and measured both final QoR and early-iteration
convergence for Sonnet~4.6 and Kimi~K2.5.}

\subsubsection{Retrieval Effects on Final QoR}

Comparing the retrieval-enabled tranches against the no-retrieval tranches in
Table~\ref{tab:benchmarks_results_merged} shows that retrieval does not help
the final \(600\)-iteration endpoint on this suite. In co-optimization, it
regresses all six circuit-level endpoints and raises the geometric-mean
objective from \(0.85477\) to \(0.85961\) for Sonnet~4.6 and from \(0.85439\)
to \(0.85948\) for Kimi~K2.5. This motivates a separate look at earlier
checkpoints to isolate whether retrieval changes convergence speed rather than
the final endpoint.

\subsubsection{Retrieval Effects on Early Convergence}

\begin{table*}[!t]
    \centering
    \scriptsize
    \setlength{\tabcolsep}{2.5pt}
    \renewcommand{\arraystretch}{1.03}

    \begin{minipage}[t]{0.49\textwidth}
    \centering
    \textbf{Sonnet~4.6}\\[0.3em]
    \resizebox{\linewidth}{!}{%
    \begin{tabular}{|c|c|cc|cc|cc|}
        \hline
        \multirow{2}{*}{\textbf{Tech}} & \multirow{2}{*}{\textbf{Circuit}}
        & \multicolumn{2}{c|}{\textbf{$WL$ Optimization}}
        & \multicolumn{2}{c|}{\textbf{$ECP$ Optimization}}
        & \multicolumn{2}{c|}{\textbf{Co-Optimization}} \\
        \cline{3-8}
        & & $WL$ & $ECP$* & $ECP$ & $WL$* & $WL$** & $ECP$** \\
        \hline \hline
        \multicolumn{8}{|c|}{\textbf{200 Iterations, 12 Parameters, No Retrieval}} \\ \hline
        \multirow{3}{*}{SKY130HD}
        & IBEX & 630094 & 12.25 & 10.07 & 722571 & 703646 & 10.09 \\
        & AES & 423774 & 4.83 & 3.81 & 539114 & 495966 & 3.89 \\
        & JPEG & 1068561 & 6.94 & 6.53 & 1293586 & 1090704 & 6.61 \\
        \hline
        \multirow{3}{*}{ASAP7}
        & IBEX & 96145 & 1278 & 1189 & 103220 & 100889 & 1193 \\
        & AES & 67021 & 457 & 431 & 74375 & 71291 & 438 \\
        & JPEG & 260970 & 904 & 877 & 276194 & 271271 & 883 \\
        \hline
        \hline \hline
        \multicolumn{8}{|c|}{\textbf{200 Iterations, 12 Parameters, Retrieval}} \\ \hline
        \multirow{3}{*}{SKY130HD}
        & IBEX & 625960 & 12.20 & 10.08 & 719783 & 699197 & 10.05 \\
        & AES & 422567 & 4.80 & 3.81 & 539633 & 496717 & 3.89 \\
        & JPEG & 1069710 & 6.93 & 6.48 & 1286385 & 1081780 & 6.58 \\
        \hline
        \multirow{3}{*}{ASAP7}
        & IBEX & 95370 & 1278 & 1189 & 102448 & 100205 & 1186 \\
        & AES & 67011 & 456 & 428 & 74148 & 71289 & 438 \\
        & JPEG & 260235 & 902 & 870 & 276015 & 269256 & 883 \\
        \hline
        \hline
    \end{tabular}%
    }
    \end{minipage}
    \hfill
    \begin{minipage}[t]{0.49\textwidth}
    \centering
    \textbf{Kimi~K2.5}\\[0.3em]
    \resizebox{\linewidth}{!}{%
    \begin{tabular}{|c|c|cc|cc|cc|}
        \hline
        \multirow{2}{*}{\textbf{Tech}} & \multirow{2}{*}{\textbf{Circuit}}
        & \multicolumn{2}{c|}{\textbf{$WL$ Optimization}}
        & \multicolumn{2}{c|}{\textbf{$ECP$ Optimization}}
        & \multicolumn{2}{c|}{\textbf{Co-Optimization}} \\
        \cline{3-8}
        & & $WL$ & $ECP$* & $ECP$ & $WL$* & $WL$** & $ECP$** \\
        \hline \hline
        \multicolumn{8}{|c|}{\textbf{200 Iterations, 12 Parameters, No Retrieval}} \\ \hline
        \multirow{3}{*}{SKY130HD}
        & IBEX & 630173 & 12.27 & 10.06 & 722538 & 704856 & 10.11 \\
        & AES & 423358 & 4.83 & 3.82 & 538305 & 496170 & 3.88 \\
        & JPEG & 1069289 & 6.95 & 6.55 & 1294602 & 1088907 & 6.60 \\
        \hline
        \multirow{3}{*}{ASAP7}
        & IBEX & 95931 & 1277 & 1192 & 103560 & 101098 & 1195 \\
        & AES & 67083 & 455 & 432 & 74631 & 71339 & 437 \\
        & JPEG & 260306 & 905 & 876 & 275801 & 272075 & 884 \\
        \hline
        \hline \hline
        \multicolumn{8}{|c|}{\textbf{200 Iterations, 12 Parameters, Retrieval}} \\ \hline
        \multirow{3}{*}{SKY130HD}
        & IBEX & 626039 & 12.23 & 10.06 & 719750 & 700407 & 10.07 \\
        & AES & 422151 & 4.81 & 3.82 & 538824 & 496921 & 3.88 \\
        & JPEG & 1070439 & 6.94 & 6.50 & 1287401 & 1079983 & 6.57 \\
        \hline
        \multirow{3}{*}{ASAP7}
        & IBEX & 95156 & 1278 & 1192 & 102788 & 100415 & 1188 \\
        & AES & 67074 & 455 & 428 & 74403 & 71337 & 436 \\
        & JPEG & 259572 & 904 & 869 & 275623 & 270061 & 885 \\
        \hline
        \hline
    \end{tabular}%
    }
    \end{minipage}

    \caption{\added{\textbf{ORFS-agent} \(200\)-iteration checkpoints under
    \textbf{Sonnet~4.6} and \textbf{Kimi~K2.5}. Each panel reports the absolute
    values without and with retrieval for the \(12\)-parameter setting. Column
    meanings follow Table~\ref{tab:benchmarks_results_merged}.}}
    \label{tab:benchmarks_results_lowiter}
\end{table*}

\begin{addedblock}
\begin{figure*}[!t]
    \centering
    \includegraphics[width=0.95\textwidth]{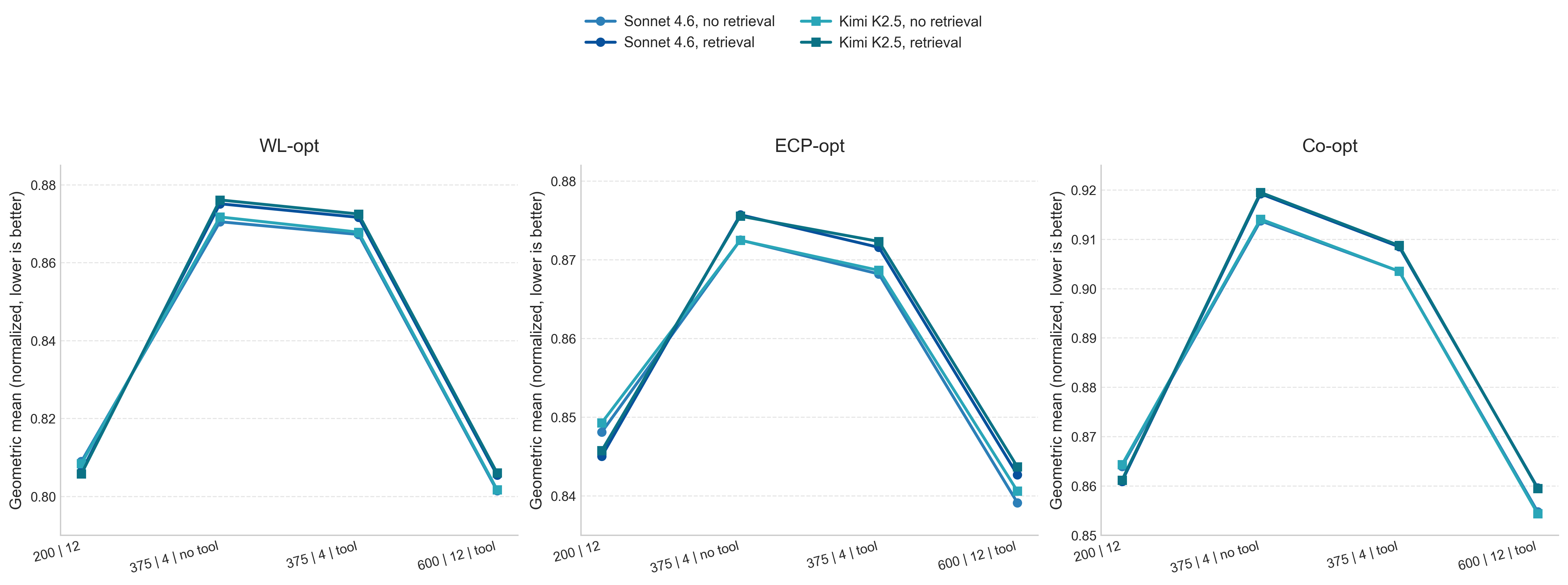}
    \caption{\added{Geometric-mean trend summary of
    Tables~\ref{tab:benchmarks_results_merged} and
    \ref{tab:benchmarks_results_lowiter}. Each panel normalizes the reported
    absolute values by the default ORFS baseline and then aggregates across the
    six circuits. The figure shows, in one view, how Sonnet~4.6 and Kimi~K2.5
    evolve from the \(200\)-iteration checkpoints to the \(375\)- and
    \(600\)-iteration settings, and how retrieval helps the low-iteration regime
    but regresses the final \(600\)-iteration endpoint.}}
    \Description{Three line plots showing geometric-mean normalized wirelength-only, effective-clock-period-only, and co-optimization objectives across four configurations: 200 iterations with 12 parameters, 375 iterations with 4 parameters and no tool use, 375 iterations with 4 parameters and tool use, and 600 iterations with 12 parameters and tool use. Each plot compares Sonnet 4.6 and Kimi K2.5 with and without retrieval.}
    \label{fig:configuration_summary}
\end{figure*}
\end{addedblock}

Table~\ref{tab:benchmarks_results_merged} shows that enabling retrieval does not
improve the 600-iteration endpoint QoR, and slightly regresses the
co-optimization objective on average. Inspecting the tool traces reveals three
empirical patterns:

\begin{itemize}
    \item \textbf{Search was front-loaded:} across 600 iterations (24 serial
    iterations with 25 parallel runs), no search query is observed after
    the eighth serial iteration.
    \item \textbf{Queries were relevant:} example queries included \texttt{CTS},
    \texttt{ASAP7 PDK}, and related OpenROAD tuning terms.
    \item \textbf{Returns diminished:} later queries often retrieved overlapping
    sources due to low query variation (e.g., \texttt{CTS} and \texttt{CTS
    wirelength}).
\end{itemize}

These patterns suggest retrieval acted primarily as early scaffolding rather than
sustained guidance. Consistent with this, Table~\ref{tab:benchmarks_results_lowiter}
shows modest early-iteration gains: at 200 iterations, retrieval reduced the
geometric-mean co-optimization objective from \(\approx 0.864\) to
\(\approx 0.861\) for both Sonnet~4.6 and Kimi~K2.5, while at 600 iterations it
increased it from \(\approx 0.855\) to \(\approx 0.860\). Figure~\ref{fig:configuration_summary}
compresses the same tables into a geometric-mean view across budgets and makes
the same pattern visually immediate: retrieval shifts the low-iteration
operating point, but the best final endpoints are reached without retrieval.

\noindent\textbf{Search trace example.} We observe that web and scholarly 
search can bias a model toward
superficially plausible conclusions by overweighting highly retrievable sources
rather than decision-relevant ones. In our probe with both \textbf{Brave}
(\texttt{web\_search}) and \textbf{OpenAlex} (\texttt{openalex\_lookup}) 
enabled, both models combined exploratory retrieval (e.g., 
querying \texttt{CTS} or PDK terms) with targeted artifact lookup 
(e.g., OpenROAD-flow-scripts \texttt{config.mk}
files and flow-variable documentation) before proposing parameter updates.
Notably, the same APIs were able to retrieve the original ORFS-agent
paper~\cite{orfsagent_mlcad25};
we do not treat this as data leakage because the optimization decisions remained
grounded in current-run metrics and tool outputs rather than static prior text.

\subsection{\added{Robustness, Sensitivity,
and Ablations}}
\label{sec:ibexablations}

\noindent\textbf{Effects of iterations and indirect optimization.} We consider the 
results of the joint $WL,ECP$ optimization as a 
function of the total serial iterations (keeping the number of parallel runs 
fixed at $25$). The results are shown in Table~\ref{tab:trajectory}, along with the
other tracked figures of merit (Area, Count, PDP, Power). Note that each row
represents a separate run that is executed up to the corresponding number of serial
iterations with $25$ parallel runs. For example, $5$ should be considered as 
a total of $25 \times 5 = 125$ OpenROAD calls. \added{Table~\ref{tab:latest_model_trajectory_abs}
later reuses the same six-point ASAP7-IBEX grid for Sonnet~4.6 and
Kimi~K2.5, and Tables~\ref{tab:reasoning_sonnet46}--\ref{tab:reasoning_search_200}
summarize the corresponding reasoning patterns.}

\begin{table}[!ht]
    \centering
    \scriptsize
    \begin{tabular}{|c|c|c|c|c|c|c|}
        \hline
        & $WL$ & $ECP$ & Area & Count & PDP & Power \\ \hline
        5 & 108267 & 1353 & 2711 & 21614 & 78.474 & 0.058 \\ \hline
        10 & 108530 & 1334 & 2701 & 21529 & 73.37 & 0.055 \\ \hline
        15 & 111557 & 1294 & 2742 & 21875 & 89.286 & 0.069 \\ \hline
        20 & 110776 & 1293 & 2735 & 21831 & 89.217 & 0.069 \\ \hline
        25 & 112495 & 1298 & 2729 & 21796 & 85.734 & 0.066 \\ \hline
        30 & 108409 & 1328 & 2711 & 21614 & 77.024 & 0.058 \\ \hline
    \end{tabular}
    \caption{Optimization trajectory with 5 to 30 iterations, no tool use, 4-parameter setting.}
    \label{tab:trajectory}
\end{table}

This phenomenon can be well understood in the light of the \textbf{strong 
inter-correlations} of the data, as discussed in Section~\ref{subsec:multi-objective}.

\noindent\textbf{Reversibility.} Table~\ref{tab:trajectory} indicates that optimizing 
wirelength and $ECP$ jointly also 
optimizes other parameters. 
We also reverse this, and analyze the effect on wirelength and $ECP$ when 
any one of these other parameters is optimized individually. The results of this process 
are presented in Table~\ref{tab:rev}. Notably, outcomes are
\emph{asymmetric} in that we see noticeably worse   $ECP$ outcomes, but with 
significantly lower power consumption.

\begin{table}[!ht]
    \centering
    \scriptsize
    \begin{tabular}{|c|c|c|c|c|c|c|}
        \hline
        & Count & Area & PDP & Power & $WL$ & $ECP$ \\ \hline
        Count & 21419 & 2679 & 68.632 & 0.046 & 109084 & 1492 \\ \hline
        Area & 21437 & 2674 & 70.952 & 0.049 & 117411 & 1448 \\ \hline
        PDP & 21432 & 2686 & 68.310 & 0.046 & 106070 & 1485 \\ \hline
        Power & 21516 & 2689 & 68.540 & 0.046 & 107350 & 1490 \\ \hline
        $WL$ & 21540 & 2701 & 76.828 & 0.058 & 106044 & 1324 \\ \hline
        $ECP$ & 21799 & 2748 & 86.483 & 0.067 & 113854 & 1281 \\ \hline
    \end{tabular}
    \caption{Results of optimizing the row-level objective on column-level 
    outcomes: ASAP7-IBEX, no tool use, and 4-parameter setting.}
    \label{tab:rev}
\end{table}

To confirm that our results are consistent across tool use choices, 
we replicate this experiment using ASAP7-IBEX with 12 parameters 
and tool use enabled; results are presented in Table~\ref{tab:revtool}.

\begin{table}[!ht]
    \centering
    \scriptsize
\begin{tabular}{|c|c|c|c|c|c|c|}
\hline
& Count & Area & PDP & Power & $WL$ & $ECP$ \\ \hline
Count & 21356 & 2683 & 75.28 & 0.054 & 112984 & 1394 \\ \hline
Area & 21382 & 2642 & 76.21 & 0.058 & 120811 & 1314 \\ \hline
PDP & 21486 & 2706 & 59.41 & 0.047 & 107822 & 1264 \\ \hline
Power & 21416 & 2683 & 59.06 & 0.046 & 104622 & 1284 \\ \hline
$WL$ & 21682 & 2672 & 99.68 & 0.078 & 97305 & 1278 \\ \hline
$ECP$ & 21674 & 2684 & 102.34 & 0.082 & 105562 & 1248 \\ \hline
\end{tabular}
    \caption{Results of optimizing the row-level objective on column-level 
    outcomes: ASAP7-IBEX, tool use, and 12-parameter setting.}
    \label{tab:revtool}
\end{table}

\begin{figure}
    \centering
	    \includegraphics[width=\columnwidth]{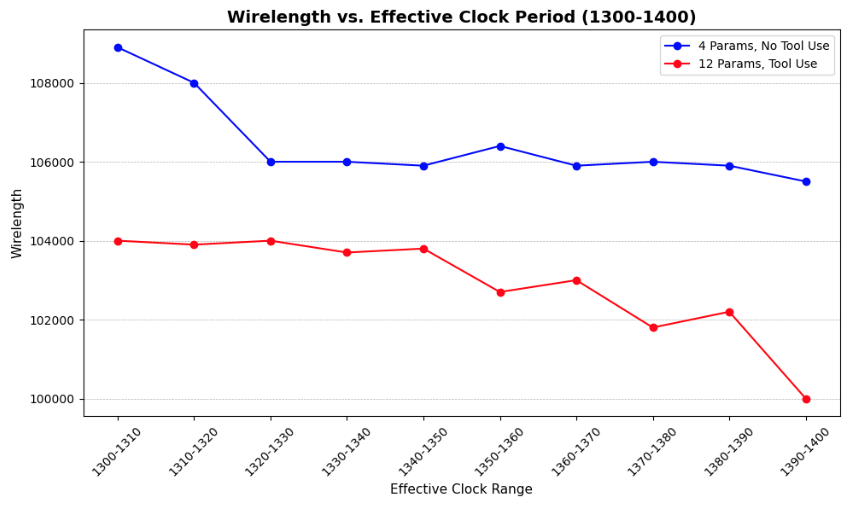}
	    \caption{Pareto frontier of $ECP$ versus $WL$ under \( 4\% \) tolerance, relative to 
	    the baseline. The blue line represents the case with 4 parameters and no tool 
	    use, and the red line represents the 12-parameter, tools-enabled setting.}
    \Description{Pareto frontier plot comparing effective clock period versus wirelength under a 4 percent constraint tolerance, with separate curves for 4-parameter/no-tool and 12-parameter/tool configurations.}
    \label{fig:pareto}
\end{figure}

The \( 4\% \) tolerance case naturally gives rise
to a Pareto frontier scatterplot of $ECP$ versus
$WL$, which is shown in Figure~\ref{fig:pareto}.

\subsubsection{\added{Data Leakage, Confidence,
and Robustness}}
\label{sec:robustness}

In this section, we analyze statistical significance in Table~\ref{tab:statsigicml},
and study prompt-level ablations in Tables~\ref{tab:promptdomain} and 
~\ref{tab:promptrole}. 
\added{Where Kimi K2.5 data are available, we report side-by-side values in the
format Sonnet~3.5\,/\,Kimi~K2.5 to enable direct cross-model comparison.}
Table~\ref{tab:promptdomain} presents outcomes of separate ablations of PDK (i.e., platform), circuit, and parameters.
We refer readers to the implementation codebase~\cite{orfsagent_code}
to understand the nature of this metadata.
The medians align with the conclusions in Sections~\ref{sec:exp} and
\ref{sec:conclusion}, and the interquartile ranges are small relative to the
medians, supporting robustness.

Table~\ref{tab:promptrole} shows additional ablations, where ``role'' refers to
the EDA-expert role that the LLM is asked to assume during prompting.

\begin{table}[ht]
  \centering
  \scriptsize
  \setlength{\tabcolsep}{3pt}
  \begin{tabular}{|c|c|c|c|c|c|c|}
    \hline
    Metric & $k\!=\!1$ & $k\!=\!5$ & $k\!=\!10$ & $k\!=\!15$ & $k\!=\!19$ & Reported \\ \hline
    $WL$    &  99820\added{\,/\,97915} &  98512\added{\,/\,96632} &  97360\added{\,/\,95502} &  96075\added{\,/\,94241} &  94980\added{\,/\,93167} &  97305\added{\,/\,95448} \\ \hline
    $ECP$*  &   1335\added{\,/\,1319} &   1301\added{\,/\,1286} &   1275\added{\,/\,1260} &   1251\added{\,/\,1236} &   1223\added{\,/\,1209} &   1278\added{\,/\,1263} \\ \hline
    $ECP$   &   1294\added{\,/\,1215} &   1271\added{\,/\,1194} &   1246\added{\,/\,1170} &   1224\added{\,/\,1149} &   1197\added{\,/\,1124} &   1248\added{\,/\,1172} \\ \hline
    $WL$*   & 108430\added{\,/\,104516} & 107180\added{\,/\,103312} & 105598\added{\,/\,101787} & 104125\added{\,/\,100367} & 102470\added{\,/\,98772} & 105562\added{\,/\,101752} \\ \hline
    $WL$**  & 105230\added{\,/\,102623} & 103720\added{\,/\,101150} & 102178\added{\,/\,99647} & 101620\added{\,/\,99102} &  99740\added{\,/\,97269} & 102205\added{\,/\,99673} \\ \hline
    $ECP$** &   1311\added{\,/\,1226} &   1283\added{\,/\,1200} &   1259\added{\,/\,1177} &   1232\added{\,/\,1152} &   1201\added{\,/\,1123} &   1260\added{\,/\,1178} \\ \hline
  \end{tabular}
  \caption{Statistical significance over 20 independent ORFS-agent runs on the
  ASAP7-IBEX circuit. Each cell shows Sonnet~3.5\added{\,/\,Kimi~K2.5} values.
  Columns \(k\!=\!1,5,10,15,19\) report the $k^{th}$ worst value among the 20
  outcomes; “Reported” is the single-run
  value reported earlier in Section~\ref{sec:single-objective}.}
  \label{tab:statsigicml}
\end{table}

\noindent
\begin{minipage}[t]{0.49\linewidth}
  \centering
  \scriptsize
  \setlength{\tabcolsep}{3pt}
  \renewcommand{\arraystretch}{1.05}
  \resizebox{\linewidth}{!}{%
  \begin{tabular}{|c|c|c|c|c|c|c|}
    \hline
    Configuration & $WL$ & $ECP$* & $ECP$ & $WL$* & $WL$** & $ECP$** \\ \hline
    Baseline          &  97305 & 1278 & 1248 & 105562 & 102205 & 1260 \\ \hline
    No~PDK            &  99874 & 1385 & 1296 & 107284 & 105210 & 1275 \\ \hline
    No~circuit        &  98743 & 1368 & 1301 & 106938 & 104782 & 1280 \\ \hline
    Both~removed      & 106044 & 1324 & 1281 & 113854 & 114576 & 1287 \\ \hline
  \end{tabular}%
  }
  \captionof{table}{Domain-knowledge ablation for ASAP7-IBEX. “No PDK” removes
  technology details, “No circuit” removes circuit details, and “Both removed”
  drops both. Larger $WL$/$ECP$ values are worse.}
  \label{tab:promptdomain}
\end{minipage}\hfill
\begin{minipage}[t]{0.49\linewidth}
  \centering
  \scriptsize
  \setlength{\tabcolsep}{3pt}
  \renewcommand{\arraystretch}{1.05}
  \resizebox{\linewidth}{!}{%
  \begin{tabular}{|c|c|c|c|c|c|c|}
    \hline
    Prompt variant & $WL$ & $ECP$* & $ECP$ & $WL$* & $WL$** & $ECP$** \\ \hline
    Baseline           & 97305\added{\,/\,95448} & 1278\added{\,/\,1263} & 1248\added{\,/\,1172} & 105562\added{\,/\,101752} & 102205\added{\,/\,99673} & 1260\added{\,/\,1178} \\ \hline
    No\,role           & 98044\added{\,/\,96173} & 1292\added{\,/\,1277} & 1255\added{\,/\,1179} & 106114\added{\,/\,102284} & 102934\added{\,/\,100384} & 1267\added{\,/\,1185} \\ \hline
    Blank\,params      & 99130\added{\,/\,97238} & 1310\added{\,/\,1295} & 1270\added{\,/\,1193} & 107039\added{\,/\,103176} & 103848\added{\,/\,101275} & 1278\added{\,/\,1195} \\ \hline
  \end{tabular}%
  }
  \captionof{table}{Prompt-sensitivity ablation for ASAP7-IBEX. Each cell shows
  Sonnet~3.5\added{\,/\,Kimi~K2.5}. “No role” deletes the system-role
  description; “Blank param” deletes parameter-guidance text.}
  \label{tab:promptrole}
\end{minipage}

Over the course of our work, the large amount of data from Anthropic naturally raises concerns 
about whether the LLM is simply caching or memorizing what it observes. This 
includes the possibility of interactions with other OpenROAD users who may be 
testing Claude models on similar outputs. We test these concerns in the following 
ways.

\textbf{Circuit obfuscation.} To assess potential memorization, we obfuscate the 
name of the ASAP7-IBEX circuit and subject it to the same iteration-versus-performance
ablation test. The circuit name is changed to \texttt{obfs}, while the design 
platform input to the \sk{ORFS-agent} remains ASAP7. Additionally, configuration settings 
are modified to rename the top-level module from ``IBEX" to \texttt{obfs}. The ORFS-agent is 
also restricted from grepping log outputs that refer to the original name or netlist 
identifiers.  Results are given in Tables~\ref{tab:obfuscation} and ~\ref{tab:obfuscationfull}.

\noindent
\begin{minipage}[t]{0.49\linewidth}
  \centering
  \scriptsize
  \setlength{\tabcolsep}{3pt}
  \renewcommand{\arraystretch}{1.05}
  \resizebox{\linewidth}{!}{%
  \begin{tabular}{|c|c|c|c|c|c|c|}
    \hline
    & $WL$ & $ECP$ & Area & Count & PDP & Power \\ \hline
    5  & 110605\added{\,/\,103104} & 1311\added{\,/\,1271} & 2731\added{\,/\,2731} & 21728\added{\,/\,21728} & 86.526\added{\,/\,83.886} & 0.066\added{\,/\,0.066} \\ \hline
    10 & 113186\added{\,/\,105510} & 1304\added{\,/\,1264} & 2741\added{\,/\,2741} & 21972\added{\,/\,21972} & 89.976\added{\,/\,87.216} & 0.069\added{\,/\,0.069} \\ \hline
    15 & 110358\added{\,/\,102874} & 1296\added{\,/\,1257} & 2735\added{\,/\,2735} & 21735\added{\,/\,21735} & 88.128\added{\,/\,85.476} & 0.068\added{\,/\,0.068} \\ \hline
    20 & 111918\added{\,/\,104328} & 1281\added{\,/\,1242} & 2746\added{\,/\,2746} & 21854\added{\,/\,21854} & 89.670\added{\,/\,86.940} & 0.070\added{\,/\,0.070} \\ \hline
    25 & 110965\added{\,/\,103440} & 1283\added{\,/\,1244} & 2733\added{\,/\,2733} & 21732\added{\,/\,21732} & 85.961\added{\,/\,83.348} & 0.067\added{\,/\,0.067} \\ \hline
    30 & 112645\added{\,/\,105006} & 1273\added{\,/\,1234} & 2746\added{\,/\,2746} & 21882\added{\,/\,21882} & 90.383\added{\,/\,87.614} & 0.071\added{\,/\,0.071} \\ \hline
  \end{tabular}%
  }
  \captionof{table}{Obfuscated ASAP7-IBEX results, no tools, 4 parameters.
  Values are Sonnet~3.5\added{\,/\,Kimi~K2.5}.}
  \label{tab:obfuscation}
\end{minipage}\hfill
\begin{minipage}[t]{0.49\linewidth}
  \centering
  \scriptsize
  \setlength{\tabcolsep}{3pt}
  \renewcommand{\arraystretch}{1.05}
  \resizebox{\linewidth}{!}{%
  \begin{tabular}{|c|c|c|c|c|c|c|}
    \hline
    & $WL$ & $ECP$ & {Area} & {Count} & {PDP} & {Power} \\
    \hline
    5  & 114861\added{\,/\,112015} & 1316\added{\,/\,1230} & 2716\added{\,/\,2716} & 22015\added{\,/\,22015} & 84.22\added{\,/\,78.72} & 0.064\added{\,/\,0.064} \\ \hline
    10 & 105788\added{\,/\,103167} & 1294\added{\,/\,1210} & 2728\added{\,/\,2728} & 21986\added{\,/\,21986} & 90.58\added{\,/\,84.70} & 0.070\added{\,/\,0.070} \\ \hline
    15 & 106724\added{\,/\,104080} & 1286\added{\,/\,1202} & 2735\added{\,/\,2735} & 21924\added{\,/\,21924} & 87.45\added{\,/\,81.74} & 0.068\added{\,/\,0.068} \\ \hline
    20 & 106245\added{\,/\,103613} & 1278\added{\,/\,1195} & 2704\added{\,/\,2704} & 21864\added{\,/\,21864} & 84.35\added{\,/\,78.87} & 0.066\added{\,/\,0.066} \\ \hline
    25 & 103871\added{\,/\,101298} & 1276\added{\,/\,1193} & 2694\added{\,/\,2694} & 21882\added{\,/\,21882} & 90.60\added{\,/\,84.70} & 0.071\added{\,/\,0.071} \\ \hline
    30 & 103244\added{\,/\,100686} & 1272\added{\,/\,1189} & 2711\added{\,/\,2711} & 21808\added{\,/\,21808} & 94.13\added{\,/\,87.99} & 0.074\added{\,/\,0.074} \\ \hline
  \end{tabular}%
  }
  \captionof{table}{Obfuscated ASAP7-IBEX results, tools, 12 parameters.
  Values are Sonnet~3.5\added{\,/\,Kimi~K2.5}.}
  \label{tab:obfuscationfull}
\end{minipage}

\textbf{Statistical significance.} We run the ASAP7-IBEX optimization for 
$ECP$, $WL$ $10$ times, with results given in Tables~\ref{tab:statsig} 
and ~\ref{tab:statsigfull}. 
Each individual row indicates the outcome with the lowest $ECP$ in the final $25$ 
iterations of an individual run.
Even the worst of the $10$ produces a result 
that is superior to OR-AutoTuner in terms of $ECP$.

\begin{table}[!ht]
    \centering
    \scriptsize
    \begin{tabular}{|c|c|c|c|c|c|c|}
        \hline
        & $WL$ & $ECP$ & Area & Count & PDP & Power \\ \hline
        Run 1 & 116479 & 1290 & 2744 & 21781 & 90.300 & 0.070 \\ \hline
        Run 2 & 110232 & 1286 & 2736 & 21775 & 90.020 & 0.070 \\ \hline
        Run 3 & 121092 & 1287 & 2778 & 22181 & 91.377 & 0.071 \\ \hline
        Run 4 & 114576 & 1287 & 2746 & 22013 & 84.942 & 0.066 \\ \hline
        Run 5 & 109107 & 1303 & 2720 & 21636 & 80.786 & 0.062 \\ \hline
        Run 6 & 111054 & 1275 & 2740 & 21816 & 91.800 & 0.072 \\ \hline
        Run 7 & 116722 & 1279 & 2758 & 22098 & 89.530 & 0.070 \\ \hline
        Run 8 & 115933 & 1281 & 2766 & 21918 & 90.951 & 0.071 \\ \hline
        Run 9 & 109230 & 1309 & 2724 & 21656 & 82.467 & 0.063 \\ \hline
        Run 10 & 109342 & 1299 & 2728 & 21734 & 88.332 & 0.068 \\ \hline
    \end{tabular}
    \caption{$10$-fold runs, ASAP7-IBEX, no tools, 4 parameters.}
    \label{tab:statsig}
\end{table}

\begin{table}[!ht]
\centering 
\scriptsize
\begin{tabular}{|c|c|c|c|c|c|c|} 
\hline & $WL$ & $ECP$ & Area & Count & PDP & Power \\ \hline 
Run 1 & 102289 & 1288 & 2726 & 21688 & 100.464 & 0.078 \\ \hline 
Run 2 & 102677 & 1285 & 2715 & 21722 & 102.800 & 0.080 \\ \hline 
Run 3 & 104200 & 1254 & 2705 & 21782 & 102.828 & 0.082 \\ \hline 
Run 4 & 103384 & 1270 & 2674 & 22092 & 109.220 & 0.086 \\ \hline 
Run 5 & 104922 & 1268 & 2678 & 21954 & 84.956 & 0.067 \\ \hline 
Run 6 & 102874 & 1269 & 2691 & 21894 & 91.368 & 0.072 \\ \hline 
Run 7 & 103982 & 1289 & 2712 & 21866 & 95.386 & 0.074 \\ \hline 
Run 8 & 103581 & 1275 & 2732 & 21895 & 96.900 & 0.076 \\ \hline 
Run 9 & 106742 & 1248 & 2725 & 21906 & 88.608 & 0.071 \\ \hline 
Run 10 & 105864 & 1252 & 2698 & 21967 & 87.640 & 0.070 \\ \hline 
\end{tabular} 
\caption{10-fold runs, ASAP7-IBEX, with tools, 12 parameters.} 
\label{tab:statsigfull} 
\end{table}

\textbf{Robustness testing.} We perform two layers of robustness testing. 
First, we add noise to every measurement: we multiply every observation $O$, at 
random, with the variable $1+X_O$, where each $X_O$ is an i.i.d. uniform random 
variable drawn from $[-0.05,0.05]$. Next, we consider the scenario where our specification of 
the surrogate has gone astray (i.e., the objective 
being optimized does not track what is desired. For this, we change all the 
CTS-level observations to be multiplied by $1+X_{CTS}$, where $X_{CTS}$ is an i.i.d.
uniform random variable drawn from $[-0.2,0.2]$. The results for ASAP7-IBEX 
are presented in Table~\ref{tab:robustness}.

\begin{table}[!ht]
    \centering
    \scriptsize
    \begin{tabular}{|c|c|c|c|c|c|c|}
        \hline
        Run     & $WL$    & CP    & Area  & Count &
        PDP     & Power \\ \hline
        \multicolumn{7}{|c|}{Overall perturbation, per objective case} \\ \hline
        $WL$      & 104427 & 1318 & 2741 & 21782 & 92.26 & 0.070 \\ \hline
        $ECP$     & 106872 & 1268 & 2708 & 21905 & 98.90 & 0.078 \\ \hline
        $WL$ \& $ECP$ & 107848 & 1274 & 2685 & 21802 & 96.82 & 0.076 \\ \hline
        \multicolumn{7}{|c|}{CTS perturbation, per objective case} \\ \hline
        $WL$ & 106898 & 1305 & 2673 & 21917 & 93.96 & 0.072 \\ \hline
        $ECP$ & 112824 & 1272 & 2711 & 21904 & 96.67 & 0.076 \\ \hline
        $WL$ \& $ECP$ & 109873 & 1281 & 2705 & 21877 & 93.51 & 0.073 \\ \hline
    \end{tabular}
    \caption{Perturbed runs, with tools, 12 parameters.}
    \label{tab:robustness}
\end{table}

\subsection{\added{Decision Process Analysis}}
\label{sec:decision_process}

For the same ASAP7-IBEX co-optimization runs already summarized in
Table~\ref{tab:thinking_model_comparison} and Figure~\ref{fig:3x3_collage}, we
next inspect checkpoint-by-checkpoint behavior and abridged reasoning summaries.
This subsection does not introduce a separate experiment; it unpacks the same
thinking-model runs. Table~\ref{tab:latest_model_trajectory_abs} reports the
Sonnet~4.6 and Kimi~K2.5 co-optimization checkpoints on the same six-point
serial-iteration grid \(\{5,10,15,20,25,30\}\) used in Table~\ref{tab:trajectory}. For Figure~\ref{fig:3x3_collage}, the
normalization baseline is the default ASAP7-IBEX ORFS result in
Table~\ref{tab:benchmarks}, namely
\(WL_\alpha = 115285\) and \(ECP_\alpha = 1361\).

\begin{table*}[!t]
    \centering
    \scriptsize
    \setlength{\tabcolsep}{4pt}
    \renewcommand{\arraystretch}{1.12}
    \resizebox{\textwidth}{!}{%
    \begin{tabular}{|c|c|c|c|c|}
        \hline
        \textbf{Serial iters} &
        \textbf{Sonnet 4.6, no search} &
        \textbf{Sonnet 4.6, search} &
        \textbf{Kimi K2.5, no search} &
        \textbf{Kimi K2.5, search} \\
        \hline
        5 &
        \makecell[c]{\((108087,1277)\)\\\((0.938,0.938)\)} &
        \makecell[c]{\((107745,1274)\)\\\((0.935,0.936)\)} &
        \makecell[c]{\((108192,1278)\)\\\((0.938,0.939)\)} &
        \makecell[c]{\((107850,1274)\)\\\((0.936,0.936)\)} \\
        \hline
        10 &
        \makecell[c]{\((100889,1193)\)\\\((0.875,0.877)\)} &
        \makecell[c]{\((100205,1186)\)\\\((0.869,0.871)\)} &
        \makecell[c]{\((101098,1195)\)\\\((0.877,0.878)\)} &
        \makecell[c]{\((100415,1188)\)\\\((0.871,0.873)\)} \\
        \hline
        15 &
        \makecell[c]{\((105856,1245)\)\\\((0.918,0.915)\)} &
        \makecell[c]{\((106897,1250)\)\\\((0.927,0.918)\)} &
        \makecell[c]{\((105659,1243)\)\\\((0.917,0.913)\)} &
        \makecell[c]{\((106700,1249)\)\\\((0.926,0.918)\)} \\
        \hline
        20 &
        \makecell[c]{\((103750,1224)\)\\\((0.900,0.899)\)} &
        \makecell[c]{\((104724,1229)\)\\\((0.908,0.903)\)} &
        \makecell[c]{\((103664,1221)\)\\\((0.899,0.897)\)} &
        \makecell[c]{\((104685,1227)\)\\\((0.908,0.902)\)} \\
        \hline
        25 &
        \makecell[c]{\((101643,1202)\)\\\((0.882,0.883)\)} &
        \makecell[c]{\((102551,1207)\)\\\((0.890,0.887)\)} &
        \makecell[c]{\((101668,1200)\)\\\((0.882,0.882)\)} &
        \makecell[c]{\((102671,1205)\)\\\((0.891,0.885)\)} \\
        \hline
        30 &
        \makecell[c]{\((99537,1181)\)\\\((0.863,0.868)\)} &
        \makecell[c]{\((100378,1186)\)\\\((0.871,0.871)\)} &
        \makecell[c]{\((99673,1178)\)\\\((0.865,0.866)\)} &
        \makecell[c]{\((100656,1183)\)\\\((0.873,0.869)\)} \\
        \hline
    \end{tabular}%
    }
    \caption{\added{Thinking-model ASAP7-IBEX co-optimization trajectory. Each cell
    reports the absolute checkpoint \((WL^{**}, ECP^{**})\) on the first line
    and the normalized checkpoint
    \((WL^{**}/WL_\alpha, ECP^{**}/ECP_\alpha)\) on the second line, with
    \(WL_\alpha = 115285\) and \(ECP_\alpha = 1361\).}}
    \label{tab:latest_model_trajectory_abs}
    \label{tab:latest_model_trajectory_norm}
\end{table*}

The corresponding reasoning summaries are abridged below. We do not reproduce
raw hidden chain-of-thought or full thinking-token traces. Instead, we provide
concise reasoning summaries that match the reported checkpoints and reflect the
observable optimization logic: comparison against baseline, interpretation of
tool/search outputs, and the parameter families favored in the next candidate
batch. The endpoint values below are exactly the ones reported in the
tables.

\noindent\textbf{Sonnet 4.6: 600 Iterations, 12 Parameters, No Search.}
Sonnet 4.6 anchored all proposals to the ASAP7-IBEX default baseline
\((115285, 1361)\) and kept the task explicitly multi-objective: candidates were
favored only when both \(WL\) and \(ECP\) moved down together. In the
12-parameter space, the run treated placement-density knobs (core utilization,
padding, and layer adjustments) and timing knobs (clock period and CTS
clustering) as coupled controls rather than chasing a pure wirelength minimum.
Once both QoRs were comfortably below the baseline, the search radius
contracted, broad exploration gave way to smaller routing-resource and CTS
adjustments, and the reported endpoint in
Table~\ref{tab:benchmarks_results_merged} reached
\(WL^{**}=99537\) and \(ECP^{**}=1181\), corresponding to \((0.863, 0.868)\)
after normalization.

\noindent\textbf{Kimi K2.5: 600 Iterations, 12 Parameters, No Search.}
Kimi K2.5 anchored on the same ASAP7-IBEX baseline and, in practice, compared
proposals more numerically than the earlier Sonnet~3.5 runs. Candidate
batches stayed schema-tight and tool-compatible, which kept the 12-parameter
loop stable while exploring coupled placement, routing, and CTS knobs. The
model favored smoother parameter step sizes when one QoR improved but the other
flattened, and the accepted region moved toward balanced improvements rather
than aggressive single-metric swings. As discussed in
Section~\ref{sec:kimi_thinking_search_notes}, Kimi can emit explicit
\texttt{reasoning\_content}; only a short decision summary was carried forward
so that trace tokens did not crowd out fresh metrics. The reported endpoint in
Table~\ref{tab:benchmarks_results_merged} was \(WL^{**}=99673\) and
\(ECP^{**}=1178\), i.e., \((0.865, 0.866)\) after normalization.

\noindent\textbf{Search-Augmented Reasoning at 200 Iterations.}
For both models, retrieval was front-loaded and most helpful before the run had
fully localized useful parameter regimes. Sonnet 4.6 targeted OpenROAD tuning
artifacts more directly, surfacing ORFS flow-variable documentation, actual
\texttt{config.mk} files, and the ORFS-agent arXiv page early in the run, with
no search call observed after the eighth serial iteration. Kimi K2.5 used a
slightly broader flow-level retrieval pattern before narrowing to ORFS
tutorials, design-specific \texttt{config.mk} artifacts, and the same
self-referential ORFS-agent arXiv hit noted in
Section~\ref{subsec:self_ref_retrieval}. In both cases,
retrieval reduced uncertainty about parameter semantics early enough to improve
the \(200\)-iteration checkpoint, but later retrieval added little new signal
and did not improve the eventual \(600\)-iteration optimum.

Taken together, Tables~\ref{tab:reasoning_sonnet46},
\ref{tab:reasoning_kimi25}, and \ref{tab:reasoning_search_200} show the same
expert-like pattern across the thinking models: the agent reads current metrics,
identifies whether the active bottleneck is primarily placement, routing, or CTS,
adjusts the corresponding parameter family, and then reduces step size as the
run moves toward a more balanced tradeoff between wirelength and effective
clock period.

\begin{table}[H]
    \centering
    \scriptsize
    \renewcommand{\arraystretch}{1.08}
    \begin{tabularx}{\columnwidth}{|l|>{\raggedright\arraybackslash}X|}
        \hline
        \textbf{Phase} & \textbf{Abridged decision summary} \\
        \hline
        Early exploration &
        The run anchored all proposals to the ASAP7-IBEX default baseline
        \((115285, 1361)\) and kept the task explicitly multi-objective:
        candidates were favored only when both \(WL\) and \(ECP\) moved down
        together. In the 12-parameter space, Sonnet 4.6 treated placement-density
        knobs (core utilization, padding, layer adjustments) and timing knobs
        (clock period, CTS clustering) as coupled controls rather than chasing a
        pure wirelength minimum. \\
        \hline
        Middle refinement &
        Once both QoRs were comfortably below the baseline, the search radius
        contracted. The model retained short decision summaries from earlier
        batches, rejected proposals that helped only one metric, and shifted from
        broad exploration toward smaller routing-resource and CTS adjustments that
        preserved the stronger placement regime. \\
        \hline
        Late exploitation &
        Final batches emphasized local exploitation near the best mixed-QoR
        region and kept fresh metrics in context instead of stale exploration
        rationales. The reported endpoint in
        Table~\ref{tab:benchmarks_results_merged} was
        \(WL^{**}=99537\) and \(ECP^{**}=1181\), corresponding to
        \((0.863, 0.868)\) after normalization by the default baseline. \\
        \hline
    \end{tabularx}
    \caption{\added{Abridged reasoning summary for the Sonnet 4.6 ASAP7-IBEX
    co-optimization run whose final endpoint appears in
    Table~\ref{tab:benchmarks_results_merged}.}}
    \label{tab:reasoning_sonnet46}
\end{table}

\begin{table}[H]
    \centering
    \scriptsize
    \renewcommand{\arraystretch}{1.08}
    \begin{tabularx}{\columnwidth}{|l|>{\raggedright\arraybackslash}X|}
        \hline
        \textbf{Phase} & \textbf{Abridged decision summary} \\
        \hline
        Early exploration &
        Kimi K2.5 anchored on the same ASAP7-IBEX baseline and, in practice,
        compared proposals more numerically than the earlier Sonnet~3.5
        runs. Candidate batches stayed schema-tight and tool-compatible, which
        kept the 12-parameter loop stable while exploring coupled placement,
        routing, and CTS knobs. \\
        \hline
        Middle refinement &
        The model favored smoother parameter step sizes when one QoR improved but
        the other flattened. Layer adjustments, padding, and CTS clustering were
        treated as joint congestion/timing controls, so the accepted region moved
        toward balanced improvements rather than aggressive single-metric swings. \\
        \hline
        Late exploitation &
        As discussed in Section~\ref{sec:kimi_thinking_search_notes}, Kimi can
        emit explicit \texttt{reasoning\_content}; only a short decision summary
        was carried forward so that trace tokens did not crowd out fresh metrics.
        The reported endpoint in Table~\ref{tab:benchmarks_results_merged} was
        \(WL^{**}=99673\) and \(ECP^{**}=1178\), i.e.,
        \((0.865, 0.866)\) after normalization. \\
        \hline
    \end{tabularx}
    \caption{\added{Abridged reasoning summary for the Kimi K2.5 ASAP7-IBEX
    co-optimization run whose final endpoint appears in
    Table~\ref{tab:benchmarks_results_merged}.}}
    \label{tab:reasoning_kimi25}
\end{table}

\begin{table*}[!t]
    \centering
    \scriptsize
    \renewcommand{\arraystretch}{1.06}
    \setlength{\tabcolsep}{4pt}
    \begin{tabularx}{\textwidth}{@{}lXXr@{}}
        \toprule
        \textbf{Model} & \textbf{Observed search pattern} & \textbf{Abridged decision summary} & \textbf{Reported endpoint} \\
        \midrule
        Sonnet 4.6 &
        Search focused early on OpenROAD tuning artifacts: ORFS flow-variable
        documentation, actual \texttt{config.mk} files, and the ORFS-agent arXiv
        page all surfaced in the first part of the run. No search call appeared
        after the eighth serial iteration. &
        Retrieval reduced uncertainty about parameter semantics early, so Sonnet
        4.6 reached a competitive region sooner than in the no-search run. The
        extra search budget helped at the \(200\)-iteration checkpoint, but later
        iterations added little new signal, which is why the best
        \(600\)-iteration endpoint still came from the no-search run. &
        \(WL^{**}=100205,\; ECP^{**}=1186\) \\
        \midrule
        Kimi K2.5 &
        Kimi K2.5 began with a broader retrieval sweep, then narrowed to ORFS
        tutorials, design-specific \texttt{config.mk} artifacts, and the same
        self-referential ORFS-agent arXiv hit noted in
        Section~\ref{subsec:self_ref_retrieval}. &
        Retrieval again helped the model localize useful parameter ranges sooner,
        improving the \(200\)-iteration checkpoint relative to the no-search
        variant. Once those parameter meanings were internalized, however,
        additional search text mostly consumed context budget instead of
        improving the eventual \(600\)-iteration endpoint. &
        \(WL^{**}=100415,\; ECP^{**}=1188\) \\
        \bottomrule
    \end{tabularx}
    \caption{\added{Abridged search-enabled reasoning summaries for the \(200\)-iteration
    ASAP7-IBEX co-optimization checkpoints reported in
    Table~\ref{tab:benchmarks_results_lowiter}.}}
    \label{tab:reasoning_search_200}
\end{table*}

\section{Conclusion and Future Directions}
\label{sec:conclusion}

In this work, we propose ORFS-agent, a plug-in open-source framework that
integrates LLMs into OpenROAD for hyperparameter optimization.
Despite using 40\% fewer iterations, ORFS-agent achieves results comparable to OR-AutoTuner.
ORFS-agent adopts a modular, provider-agnostic framework, which makes it user-friendly and easy to adopt.
ORFS-agent can switch seamlessly among LLM providers (e.g., Anthropic, OpenAI, Gemini)
without reconfiguration.
Additionally, ORFS-agent avoids the cost and complexity of fine-tuning and deployment,
and requires no retraining or infrastructure changes when a new LLM is released.
Thus, our ORFS-agent can instantly benefit from model breakthroughs.
\added{Our evaluation with Claude Sonnet 4.6 and Moonshot Kimi K2.5
confirms that ORFS-agent’s gains persist as model capabilities evolve: the stronger
backends preserve or improve QoR relative to the Sonnet~3.5 baseline reported in
\cite{orfsagent_mlcad25} and
can match or modestly exceed OR-AutoTuner on average while maintaining reduced
iteration budgets.}
Unlike fine-tuning-based approaches, ORFS-agent can rapidly and cost-effectively adopt stronger models, making it both competitive and appealing.

Our future work will explore the following directions.
\added{(i)} Our current framework uses a fixed set of pre-defined tools.
In future work, we will enable the LLM agent to generate its own tools dynamically,
which avoids the hand-coding of specific tools and increases flexibility in handling different circuits and technology nodes.
(ii) \added{We plan to integrate ORFS-agent with visual feedback that lets it reason across different modalities than mere text.}
And \added{(iii)} we will evaluate ORFS-agent on larger-scale circuits and with more hyperparameters to assess its scalability and robustness. In combination with open-sourcing and OpenROAD integration, we believe that
this work will add to the foundations for new research on LLMs for physical design
and EDA.

\section*{Acknowledgments} This work is partially supported by the Samsung AI Center.

\end{document}